\documentclass{article}
\usepackage{ano2027_conference,times}

\usepackage{amsmath,amsfonts,bm}

\def\eqref#1{equation~\ref{#1}}

\def\1{\bm{1}}

\DeclareMathAlphabet{\mathsfit}{\encodingdefault}{\sfdefault}{m}{sl}
\SetMathAlphabet{\mathsfit}{bold}{\encodingdefault}{\sfdefault}{bx}{n}

\usepackage{hyperref}
\usepackage{url}
\usepackage{booktabs}
\usepackage{amssymb}
\usepackage{multirow}
\usepackage{graphicx} 
\usepackage{subcaption}
\usepackage{tabularx}
\title{From World Models to World Action Models: Rethinking Next-State Prediction}
\anofinalcopy

\renewcommand{\headrulewidth}{0pt}

\author{
\makebox[\textwidth][c]{%
\begin{minipage}{0.95\textwidth}
\centering
Tingyu Yuan$^{1,2}$, Ziming Ji$^{3,7}$, Biaoliang Guan$^{4}$, Wen Ye$^{1,2}$, Wenrui Tian$^{5}$, 
Zhaopeng Gu$^{1,2}$,
Feihong Zhang$^{6}$, Xu Yang$^{1}$, Yan Huang$^{1,2}$, Zhaowen Li$^{7,\dagger}$, Chaoyang Zhao$^{1,\dagger}$, Jinqiao Wang$^{1,2,\dagger}$ \\
    $^{1}$CASIA \quad
    $^{2}$UCAS  \quad
    $^{3}$BUPT \quad
    $^{4}$XJTU \quad
    $^{5}$WHU \quad
    $^{6}$THU \quad
    $^{7}$Yinwang Intelligent Technology Co. Ltd. \quad
\end{minipage}}
}

\begin{document}

\maketitle

\begin{abstract}
Predicting the next state is a core paradigm of World Models for modeling physical dynamics, emphasizing prediction fidelity. As World Models evolve into World-Action Models (WAMs), existing methods still fix the next state before training as RGB, a single latent feature, or a static combination of predefined targets, thereby constraining action learning to the inductive biases preserved by a particular representation. To address this limitation, we propose CF-WAM, a dynamic next-state prediction framework that samples visual, semantic, geometric, and interaction projections of the same future, standardizes them into a common video form, and supervises a unified WAM across these projections. The action-relevant constraints exposed by these projections accumulate across training steps, forcing WAM to capture the underlying state-transition structure that supports multiple projections of the same action-conditioned future. This dynamic mechanism also provides a natural cross-embodiment dynamics reference frame for Human and Robot learning. By jointly learning across different next-state parameterizations, heterogeneous Human and Robot experience can bypass appearance differences and directly contribute to shared state-transition learning, improving cross-embodiment generalization. Experiments show that CF-WAM improves both training efficiency and final control performance, while translating Human experience effectively into policy gains. 
CF-WAM achieves state-of-the-art performance on RoboCasa-GR1 with an average success rate of \(82.50\%\), while reaching \(82.65\%\) on LIBERO-Plus and up to \(84.00\%\) in real-world evaluations.

\end{abstract}

\section{Introduction}
Learning to predict the next state is a core paradigm of World Models (WMs) \citep{ha2018world}, which model physical dynamics by learning how the environment evolves \citep{hafner2019learning}.
World-Action Models (WAMs) further couple this predictive capability with action generation, enabling policies to act based on anticipated futures \citep{hafner2019dream}. However, the role of the future state changes from WM to WAM. In WMs, it primarily serves faithful prediction of future evolution. In WAMs, it must also emphasize dynamic factors that can be changed by actions or directly influence action selection. Therefore, a state primarily driven by future-prediction fidelity is not necessarily the state best suited for action learning. This raises a more fundamental question: when future prediction begins to serve action, what should the next state represent?

In most visual WMs and WAMs, the next state is predefined as future RGB observations \citep{wu2024unleashing}, a single latent feature \citep{zhou2024dino}, or a static combination of multiple targets. Yet this choice determines which future information is retained, discarded, and ultimately exploited for action learning. Visual states capture complete observable outcomes but retain substantial visual detail weakly relevant to action; semantic states \citep{nair2022r3m} emphasize entities and task structure but weaken spatial relations; geometric states \citep{huang2026pointworld} capture \(3\mathrm{D}\) layout but miss object identity and interaction progress; interaction states \citep{yuan2024general} capture contact and state changes but miss global context. Different future states induce different action-relevant inductive biases, and a state effective for one task may not remain optimal for another. Some concurrent methods introduce structured supervision by statically combining multiple future targets, but this requires the model to maintain multiple prediction spaces at every step, increasing model capacity and computation while forcing modality-specific features to compete for the limited attention capacity used for action generation, thereby diluting truly action-relevant state changes \citep{huang2022modality}.

\begin{figure}[t]
    \centering
    \includegraphics[width=0.95\linewidth]{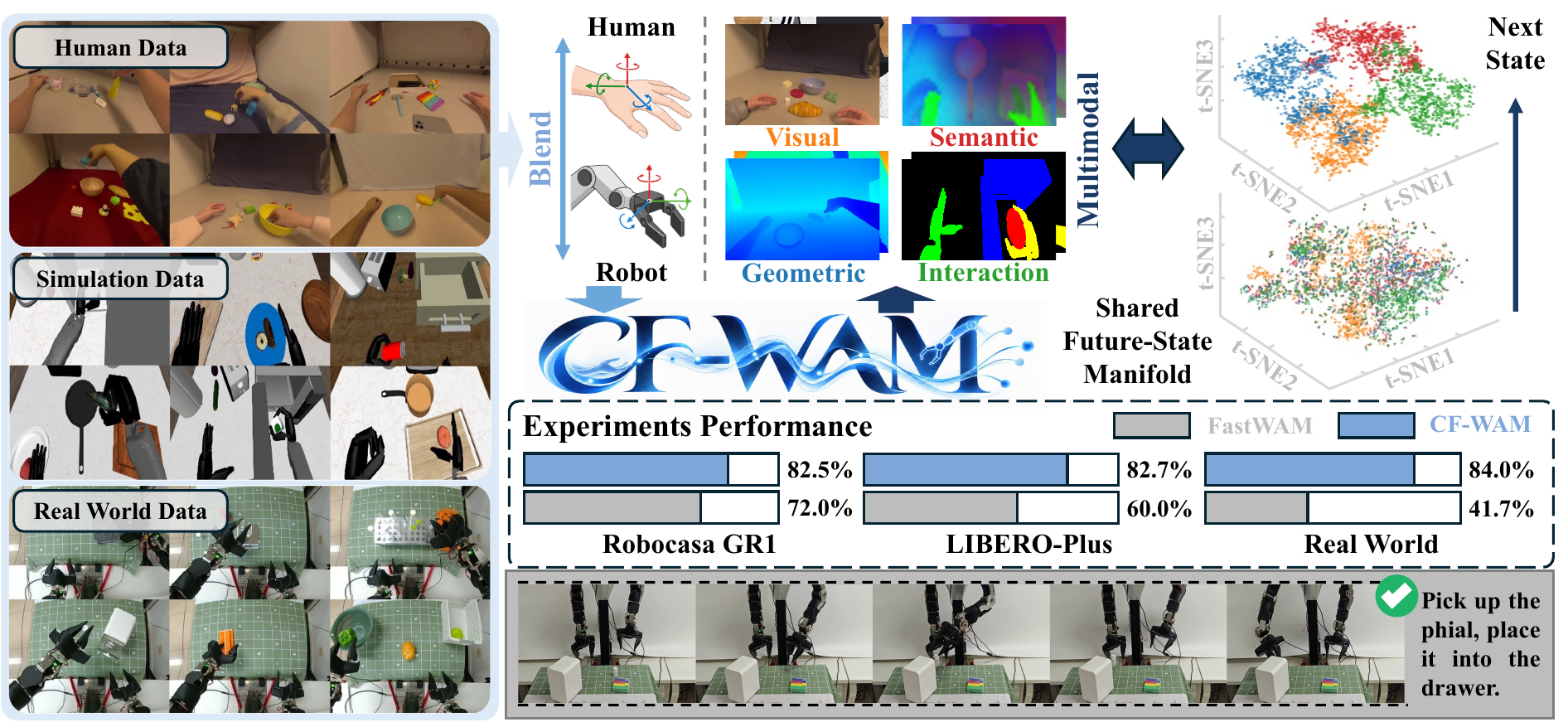}
    \caption{Overview of CF-WAM. Human and Robot data supervise multiple future-state projections of a shared underlying manifold, enabling effective Human–Robot learning and strong performance.}
    \label{fig:teaser}
\end{figure}

Meanwhile, as human manipulation data \citep{grauman2022ego4d} becomes an important source of scalable experience \citep{punamiya2026egoverse}, the choice of future representation becomes even more critical. Human data captures rich object interactions \citep{hoque2026egodex} and state changes, yet Human and Robot data differ substantially in viewpoint \citep{zhou2025mitigating}, appearance, embodiment, and action space. The same manipulation intent may produce different visual changes and motion patterns while sharing similar semantics, geometric constraints, and interaction transitions \citep{xu2024flow}. A predefined future representation can therefore be dominated by embodiment-specific appearance variations, failing to expose transferable action-relevant dynamics across the two domains. This helps explain why existing approaches often rely on Human pre-training followed by Robot post-training, or even paired Human--Robot data, limiting the effective use of Human data. The state problem is thus further amplified: effective joint learning requires a next-state formulation that bridges appearance differences \citep{wang2026humanego} while preserving action-relevant changes.

Drawing on the Platonic Representation Hypothesis, we view visual, semantic, geometric, and interaction states as different projections of the same action-conditioned future. From this perspective, the next state should not be a predefined supervision modality, but a dynamic description of the same world transition across different levels of abstraction. Based on this formulation, we propose Confluent Foresight WAM (CF-WAM), a dynamic next-state prediction framework. As illustrated in Fig.~\ref{fig:teaser}, CF-WAM expresses these projections in a common video form through a single VAE, while a unified WAM predicts the same future transition under varying coordinate systems.
Representation choice is thus transformed into a dynamic state query: what changes is the projection of the future, while the underlying action-conditioned world dynamics remain invariant. In other words, we fix the model that learns world dynamics, not the coordinate system used to describe the future state.
Furthermore, dynamic next-state prediction naturally supports heterogeneous Human--Robot learning by supervising shared world-state transitions rather than directly matching appearance or embodiment-specific action spaces. The resulting predictive representation preserves the information required for action generation while exploiting the diversity of scenes, objects, and interactions in Human data, enabling more effective data scaling and cross-embodiment transfer.

We evaluate CF-WAM across simulation, OOD, and real-robot settings. Under matched training budgets, dynamic future-state supervision improves both training efficiency and control performance over fixed state prediction, while making Human experience more effective for Robot policy learning. CF-WAM achieves an \(82.5\%\) success rate on RoboCasa-GR1, establishing state-of-the-art (SOTA) performance. Further scaling experiments show that the gains arise from learning action-relevant transition regularities across future-state projections, rather than simply adding more prediction targets. Upon acceptance, we will release the code and data.
Our main contributions are:

1. We revisit the definition of the next state in WAMs, reformulating it from a predefined representation into different projections of the same action-conditioned future, and turning representation choice from a fixed design decision into a dynamic training variable.

2. We propose CF-WAM, a dynamic next-state prediction framework that models RGB, semantic, geometric, and interaction states as complementary parameterizations of the same state transition, allowing their distinct action-relevant inductive biases to jointly shape world-action learning.

3. We show that CF-WAM relates Human and Robot data at the level of world-state transitions, enabling more effective heterogeneous data scaling and learning representations transferable across embodiments, while supporting action generation toward task goals.

4. Extensive experiments validate the effectiveness and generalization of these principles, with CF-WAM achieving SOTA performance on RoboCasa-GR1 and across multiple evaluation settings.

\section{Related Work}
\label{gen_inst}
\textbf{WMs and WAMs.}
Visual WMs primarily model environment dynamics \citep{c} by predicting future video observations, with works such as UniSim \citep{yang2023learning}, Genie \citep{bruce2024genie}, and IRASim \citep{zhu2024irasim} demonstrating generative modeling of interactive futures.
Recent WAMs further couple future prediction with action generation, including VideoVLA \citep{shen2026videovla}, Motus \citep{bi2026motus}, DreamZero \citep{ye2026world}, and Fast-WAM \citep{yuan2026fast}. These works establish future prediction as an effective signal for action learning \citep{tian2025predictive}.

\textbf{Predefined Future State for Action Learning.}
WAMs typically predefine the future-state representation before training, ranging from video to structured multimodal supervision. EgoWAM \citep{li2026egowam} compares Pixel, DINO, and \(3\mathrm{D}\) Flow targets, but still uses a single fixed future-state prediction head. SA-WAM \citep{gonzalez2026spatially}, GIFT \citep{zheng2026gift}, and Flex-\(\pi\) \citep{yan2026flex} introduce structured future supervision \citep{zhang2026dreamvla} such as affordance, DINO semantics \citep{oquab2023dinov2}, and \(3\mathrm{D}\) pointmaps, yet their predefined modalities rely on fixed multi-head or multi-stream designs. 
Such designs require the model to maintain multiple prediction spaces at every step, increasing computation while forcing representation-specific features to compete for limited action-conditioning capacity. In contrast, CF-WAM treats different representations as related projections of the same future and dynamically varies the projection choice, allowing action-relevant constraints from different coordinate systems to accumulate across training steps.

\textbf{Learning from Human Experience.}
Approaches to scaling robot learning with Human experience generally follow three paradigms. EgoScale \citep{zheng2026egoscale} adopts Human pre-training followed by Robot post-training, but staged transfer limits direct joint scaling. 
Secondly, EgoEngine \citep{liu2026egoengine} and Ego2Robot \citep{wang2026ego2robot} convert Human videos into robot-format data through retargeting or visual replacement, making performance sensitive to conversion quality. Other methods establish Human--Robot correspondence through paired demonstrations \citep{kareer2025egomimic}, shared action spaces, or explicit alignment \citep{xie2026human2robot}, but may limit scalability. In contrast, CF-WAM avoids staged transfer, data conversion, and sample-level pairing by directly co-training Human and Robot experience at the world-transition level through dynamic future-state parameterization, while shared EE supervision captures transferable motion priors.

\section{Method}
\label{headings}

\begin{figure}[t]
    \centering
    \includegraphics[width=\linewidth]{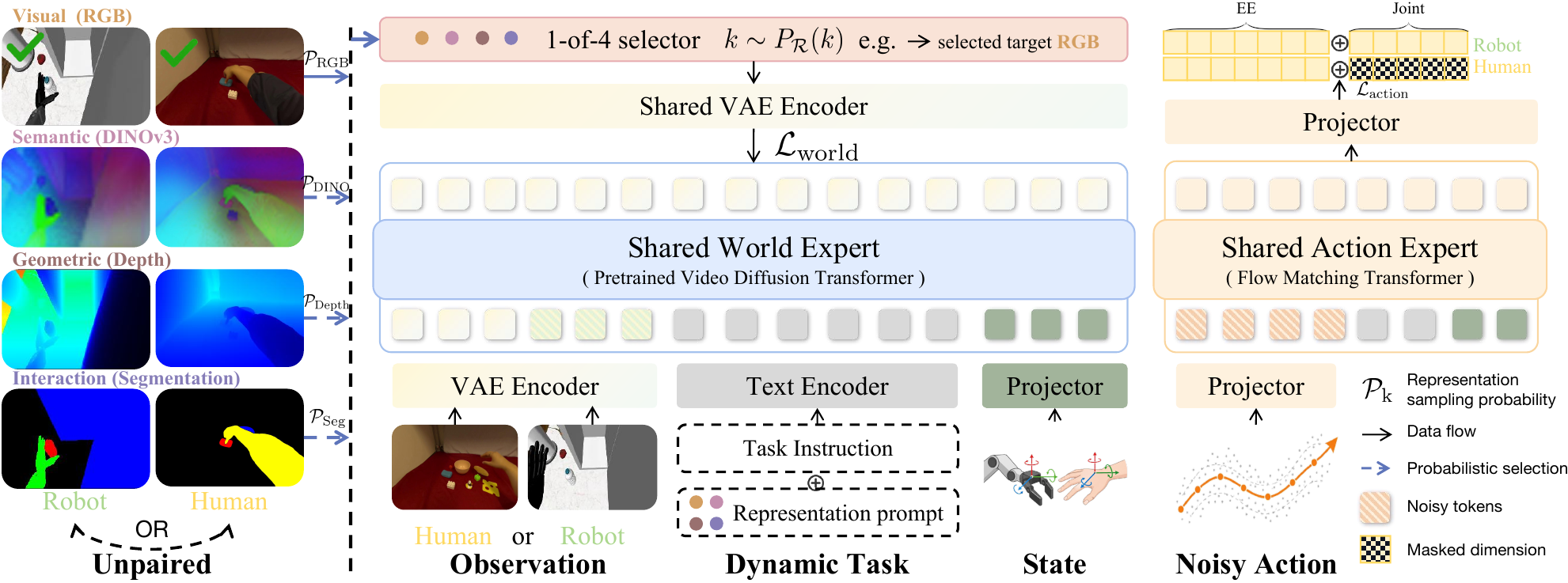}
    \caption{Overview of CF-WAM. CF-WAM dynamically samples multiple future-state projections to jointly model world dynamics and actions from unpaired Human and Robot experience.}
    \label{fig:framework}
\end{figure}

Figure~\ref{fig:framework} provides an overview of CF-WAM. Given unpaired Human and Robot experience, CF-WAM dynamically samples different future-state projections to supervise world prediction, while jointly learning action generation through a coupled world-action model. 

\subsection{Methodology}

The core idea is that a WAM should not predefine the future-state representation, but dynamically select different projections and coordinate systems of the same underlying future manifold \(s_{t+\Delta}\in\mathcal{M}\) during training.
Different future representations can be written as \(y_{t+\Delta}^{k}=\phi_k(s_{t+\Delta})\), where \(\phi_k\) denotes distinct parameterizations. They describe the same physical state transition while preserving different action-relevant information. CF-WAM follows three principles.

\textbf{Dynamic Future-State Parameterization.} We view visual, semantic, geometric, and interaction states as different projections of the same underlying future-state manifold. Rather than forming independent prediction tasks, these projections are coupled through the same underlying future and expose different action-relevant information. 
As qualitatively visualized by t-SNE in Fig.~\ref{fig:teaser}, different future-state projections exhibit a shared cross-projection structure in intermediate representations and progressively specialize into projection-specific representations toward next-state prediction, consistent with different parameterizations of the same underlying physical future.
From this perspective, CF-WAM treats projection choice itself as a dynamic training variable. A stochastic multi-task sampling strategy provides a simple realization of this formulation \citep{sundararaman2021learning}: each training sample selects one future-state projection, changing the coordinate system used to supervise the same future transition rather than the prediction problem itself. The complementary constraints from different projections thus accumulate across training steps and jointly shape the same WAM. In other words, we fix the model that learns world dynamics, not the coordinate system used to describe the future state.

\textbf{Generation as a Common Modeling Interface.} Different future-state projections do not require representation-specific architectures. CF-WAM formulates different future states as video-form generative targets and models them with a shared VAE and WAM, without introducing other separate prediction branches. Structured future understanding is thus recast as generation in different future-state forms: diversity lies in the supervision, while unity remains in the generative model. This design places representation diversity on the supervision side while sharing model structure, allowing CF-WAM to absorb multiple future-state inductive biases with almost no additional parameters.

\textbf{Unpaired Human--Robot World-Action Learning.} Dynamic future-state prediction provides Human and Robot experience with shared transition supervision. Although the two domains differ in appearance, viewpoint, embodiment, and action space, different future-state parameterizations reveal cross-embodiment regularities in semantics, spatial constraints, and interactions. This enables joint learning of world dynamics without RGB- or trajectory-level correspondence.

In summary, CF-WAM follows a simple principle: unify the model that accommodates diverse future states, rather than unifying the future states themselves; dynamically vary how the future is represented, rather than fixing a single representation before training.

\subsection{World-Action Modeling}
CF-WAM builds on a Mixture-of-Transformers (MoT) architecture with separately parameterized world and action experts. The world expert models future states with a pretrained video diffusion transformer, while the action expert predicts action trajectories. At each layer, both experts independently compute their queries, keys, and values, which are concatenated along the sequence dimension for mixed attention under an asymmetric mask \(M\):
\begin{equation}
Q=[Q^w;Q^a], \qquad
K=[K^w;K^a], \qquad
V=[V^w;V^a], \qquad
H=\operatorname{Attn}(Q,K,V;M).
\label{eq:mixed_attention}
\end{equation}

After joint attention, the hidden states are split and returned to their respective experts for further computation. The world and action experts retain independent parameters while exchanging information through layer-wise attention. This allows action prediction to exploit the current state and future dynamics encoded in world tokens, while world prediction remains independently modeled.

Given the current RGB observation \(o_t\), language instruction \(c\), and future-state target \(\tilde{y}=[o_t^{\mathrm{RGB}},y_{t+1:t+H}]\), we encode the target with the VAE as \(z_0=E_{\mathrm{VAE}}(\tilde{y})\). Using \(\sigma\in[0,1]\) as the flow-time variable, the world and action branches follow independent linear flow-matching paths:
\begin{equation}
\begin{aligned}
z_\tau = (1-\sigma_\tau)z_0 + \sigma_\tau \epsilon_z, \quad
a_\tau = (1-\sigma_\tau^a)a_0 + \sigma_\tau^a \epsilon_a.
\end{aligned}
\label{eq:flow_process}
\end{equation}

The corresponding velocity targets are \(v_z^\ast=\epsilon_z-z_0\) and \(v_a^\ast=\epsilon_a-a_0\).
The two branches use independent noise levels and schedulers, and are coupled only through MoT mixed attention.
The model jointly optimizes future prediction and action generation:
\begin{equation}
\mathcal{L}_{\mathrm{WAM}}
=
\lambda_w \mathcal{L}_{\mathrm{world}}
+
\lambda_a \mathcal{L}_{\mathrm{action}}.
\label{eq:wam_objective}
\end{equation}

Implementation details are in Appendix~A.1. This formulation assumes that the future-state target is predefined. We next make the future-state representation a dynamic training variable.

\subsection{Dynamic Future-State Prediction}
We characterize the same future transition through 4 complementary projections: visual, semantic, geometric, and interaction. Visual states capture appearance changes; semantic states emphasize object identity and scene structure; geometric states model spatial relations among objects and their changes; and interaction states highlight agents, target objects, and their evolving interactions.

In practice, we instantiate the four projections as RGB videos, DINOv3-derived semantic videos, depth videos, and hand--object segmentation videos, denoted by \(y^{\mathrm{Vis}}\), \(y^{\mathrm{Sem}}\), \(y^{\mathrm{Geo}}\), and \(y^{\mathrm{Int}}\), respectively. We define \(\mathcal{R}=\{\mathrm{Vis},\mathrm{Sem},\mathrm{Geo},\mathrm{Int}\}\) as the candidate future-state projection set. 
These representations expose complementary future information while sharing the same training form: all projections share a common video-form supervision and constitute the coordinate set dynamically sampled during training. Data construction is detailed in Appendix~A.2.

\subsubsection{Dynamic Projection Sampling}
Unlike fixing the future-state representation before training, CF-WAM treats projection choice as a dynamic training variable. For each training sample, we draw \(k\sim p_{\mathcal R}(k)\) from the projection set \(\mathcal R\). Here, \(k\) does not denote an independent prediction task, but selects the coordinate system used to describe the same underlying future transition.
Given a sampled projection \(k\), we construct the video-form target \(\tilde y^k=[o_t^{\mathrm{RGB}},\,y_{t+1:t+H}^k]\). All projections share the same current RGB observation \(o_t^{\mathrm{RGB}}\), while \(k\) determines the subsequent future sequence \(y_{t+1:t+H}^k\). For \(k\neq\mathrm{Vis}\), the first frame is replaced with the strictly time-aligned RGB frame, ensuring that all projections start from the identical current world state while only the representation of the next state changes.

Thus, each training sample is supervised by only one future-state coordinate rather than all representations simultaneously. As training proceeds, \(k\) is repeatedly resampled across samples, allowing the same world transition to be observed and constrained under different coordinate systems. The representation coordinate of the future changes, while the underlying transition remains invariant.

\subsubsection{Projection-Conditioned Generation}
Because each training sample may use a different future-state coordinate, the model must explicitly know the sampled projection. We therefore append a projection-specific condition \(c_{\mathrm{rep}}^k\) to the original task instruction \(c_{\mathrm{task}}\), yielding \(c^k=c_{\mathrm{task}}\oplus c_{\mathrm{rep}}^k\).
Here, \(c_{\mathrm{task}}\) specifies what action to perform, while \(c_{\mathrm{rep}}^k\) specifies the coordinate system used to represent future evolution. Both are processed through the model's existing language-conditioning pathway, requiring no modality-specific routing module or separate representation branch. This allows the same generative model to predict the same action-conditioned transition under different future-state coordinate systems according to \(c^k\).

\subsubsection{Dynamic Generative Learning}
For a sampled projection \(k\), the corresponding target \(\tilde y^k\) is encoded by the same VAE as \(z_0^k=E_{\mathrm{VAE}}(\tilde y^k)\), and then learned by the same world and action experts using the flow-matching objective in Sec.~3.2. Let \(\pi_k=p_{\mathcal R}(k)\), with \(\sum_{k\in\mathcal R}\pi_k=1\). At each training update, the sampled \(k\) jointly determines the future-state target \(\tilde y^k\) and its condition \(c^k\), while the model parameters remain unchanged across projections.
From a conditional-generation perspective, CF-WAM learns
\begin{equation}
p_\theta\!\left(
y_{t+1:t+H}^{k},
a_{t:t+H_a}
\mid
o_t^{\mathrm{RGB}},
c^{k}
\right),
\qquad
k\sim p_{\mathcal R}(k).
\label{eq:conditional_generation}
\end{equation}

Under different \(k\), the same parameterized model \(\theta\) captures conditional distributions of the same action-conditioned future transition in different coordinate systems.
The dynamic objective is
\begin{equation}
\mathcal{L}_{\mathrm{dyn}}
=
\mathbb{E}_{k\sim p_{\mathcal R}(k)}
\mathbb{E}_{\mathcal D_k}
\left[
\lambda_w
\mathcal{L}_{\mathrm{world}}
\!\left(
E_{\mathrm{VAE}}(\tilde y^k),c^k
\right)
+
\lambda_a
\mathcal{L}_{\mathrm{action}}
\!\left(
a\mid o_t^{\mathrm{RGB}},c^k
\right)
\right].
\label{eq:dynamic_objective}
\end{equation}

The key is not to add more prediction targets at each step, but to continuously vary the coordinate system used to supervise the same future transition. Each step uses only one projection, while the action-relevant constraints exposed by different projections accumulate across parameter updates and jointly shape the same WAM. In other words, CF-WAM keeps the world-dynamics model fixed while dynamically changing the coordinate system used to describe the future state.

\subsection{Unpaired Human–Robot World-Action Learning}
\subsubsection{Dynamic Future States as Shared Transition Supervision}
Dynamic future-state prediction provides a common transition-level interface for heterogeneous Human and Robot experience. Despite differences in appearance, viewpoint, embodiment, and action space, semantic, geometric, and interaction projections expose transferable object, spatial, and interaction changes. Human and Robot trajectories can therefore jointly supervise world dynamics without RGB- or trajectory-level correspondence or sample-level pairing.

Let the data source be \(d\in\mathcal{D}=\{\mathrm{Human},\mathrm{Robot}\}\). Human data consists of first-person manipulation, while Robot data consists of robot trajectories, with no pairing required at the episode, task, or temporal level. Together with the projection variable \(k\in\mathcal{R}\), each training sample is drawn as
\begin{equation}
(d,k)\sim p_{\mathcal{D},\mathcal{R}}(d,k)
=
p_{\mathcal{D}}(d)\,p_{\mathcal{R}}(k\mid d).
\label{eq:source_rep_sampling}
\end{equation}

Here, \(d\) determines where the experience comes from, while \(k\) determines the coordinate system in which its future transition is supervised. CF-WAM therefore varies both the experience source and future-state coordinate during training, while all samples optimize the same WAM.

\subsubsection{Cross-Embodiment EE Supervision with Robot-Specific Control}
On the action side, we decompose the action space into a cross-embodiment end-effector component and a Robot-specific joint component, \(a=[a_{\mathrm{EE}}, a_{\mathrm J}]\). 
Human wrist motion and Robot end-effector motion are represented in the same end-effector action space, allowing Human experience to directly supervise \(a_{\mathrm{EE}}\), while \(a_{\mathrm J}\) is supervised only by Robot trajectories to preserve embodiment-specific low-level control.

For Human samples, missing Robot-specific joint dimensions are masked out rather than set to zero. To distinguish it from the attention mask \(M\), we define a source-dependent action mask \(M_a^d\):
\begin{equation}
M_a^{d}
=
\begin{bmatrix}
I_{D_{\mathrm{EE}}} & 0 \\
0 & \1[d=\mathrm{R}] I_{D_{\mathrm J}}
\end{bmatrix},
\qquad
\mathcal{L}_{\mathrm{action}}^{d}
=
\mathbb{E}\!\left[
\frac{\left\|M_a^{d}r_a^{d}\right\|_2^2}
{\operatorname{tr}(M_a^{d})}
\right],
\quad
d\in\{\mathrm H,\mathrm R\}.
\label{eq:source_action_mask}
\end{equation}

Here, \(r_a^d\) denotes the action flow-matching residual in Sec.~3.2. For Human samples, \(M_a^d\) retains only the EE subspace; for Robot samples, both EE and joint subspaces are optimized. Human data can therefore provide transferable task-space motion priors without introducing spurious supervision on missing Robot-specific controls.

Combining experience-source and projection sampling, Human--Robot joint learning is written as
\begin{equation}
\mathcal{L}_{\mathrm{HR}}
=
\mathbb{E}_{(d,k)\sim p_{\mathcal D,\mathcal R}(d,k)}
\mathbb{E}_{\mathcal D_{d,k}}
\left[
\lambda_w
\mathcal{L}_{\mathrm{world}}
\!\left(
E_{\mathrm{VAE}}(\tilde y^k),c^k
\right)
+
\lambda_a
\mathcal{L}_{\mathrm{action}}^{d}
\right].
\label{eq:human_robot_objective}
\end{equation}

Here, \(\mathcal D_{d,k}\) denotes the training distribution determined by data source \(d\) and future-state projection \(k\). CF-WAM thus captures cross-embodiment transition regularities through dynamic future-state projections in the state space, while absorbing Human motion priors through EE supervision and preserving Robot-specific joint control in the action space.

\section{Experiments}
\label{others}
\subsection{Experimental Setup}
We adopt RoboCasa-GR1 as our primary simulation benchmark. It contains 24 humanoid tabletop manipulation tasks performed with the dual arms and dexterous hands of the Fourier GR-1, covering object transport and articulated-object interactions. We train on 4 nodes with 8 NVIDIA B200 GPUs per node, using a global batch size of \(32\), \(100\mathrm{K}\) training steps, and a learning rate of \(1\times10^{-4}\).
To evaluate OOD generalization, we train on LIBERO and directly test on LIBERO-Plus without any adaptation to the test environments. Training is conducted on a single NVIDIA B200 node for \(80\mathrm{K}\) steps with a learning rate of \(1\times10^{-4}\). This setting also allows us to analyze how different future-state representations generalize across tasks under distribution shifts.

For Human experience, we use the EgoDex dataset and select its pick-and-place subset as Human manipulation data. These samples are directly co-trained with Robot trajectories without requiring episode-, task-, or temporally paired Human--Robot data. We likewise construct the corresponding RGB, semantic, geometric, and interaction future-state supervision offline.

On RoboCasa-GR1, baselines include CoRE-VLA \citep{zhang2026core}, ABot-M0 \citep{yang2026abot}, JoyAI-RA 0.1 \citep{zhang2026joyai}, DIAL \citep{chen2026dial}, ACE-Ego-0 \citep{li2026ace}, DiT4DiT \citep{ma2026dit4dit}, LDA-1B \citep{lyu2026lda}, FastWAM \citep{yuan2026fast}, and WALA \citep{liu2026wala}. On LIBERO-Plus, baselines include GWM-VLA \citep{zhao2026gwm}, FoMoVLA \citep{li2026fomovla}, FastWAM and GaussianWAM \citep{zhang2026gaussianwam}. For real-world evaluation, we include \(\pi_{0.5}\) \citep{intelligence2025pi_}, GR00T-N1.6, FastWAM, and LingBot-VA \citep{li2026causal}.

\begin{figure}[t]
    \centering

    \begin{minipage}[t]{0.52\linewidth}
        \centering
        \includegraphics[width=\linewidth]{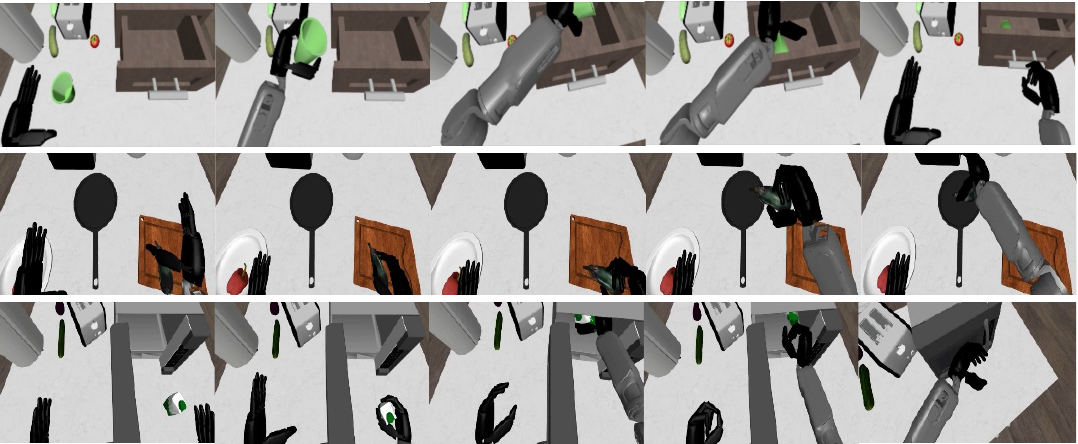}
        \captionof{figure}{
        Representative closed-loop manipulation trajectories of CF-WAM on RoboCasa-GR1.
        }
        \label{fig:robocasa_rollouts}
    \end{minipage}
    \hfill
    \begin{minipage}[t]{0.47\linewidth}
        \centering
        \includegraphics[width=\linewidth]{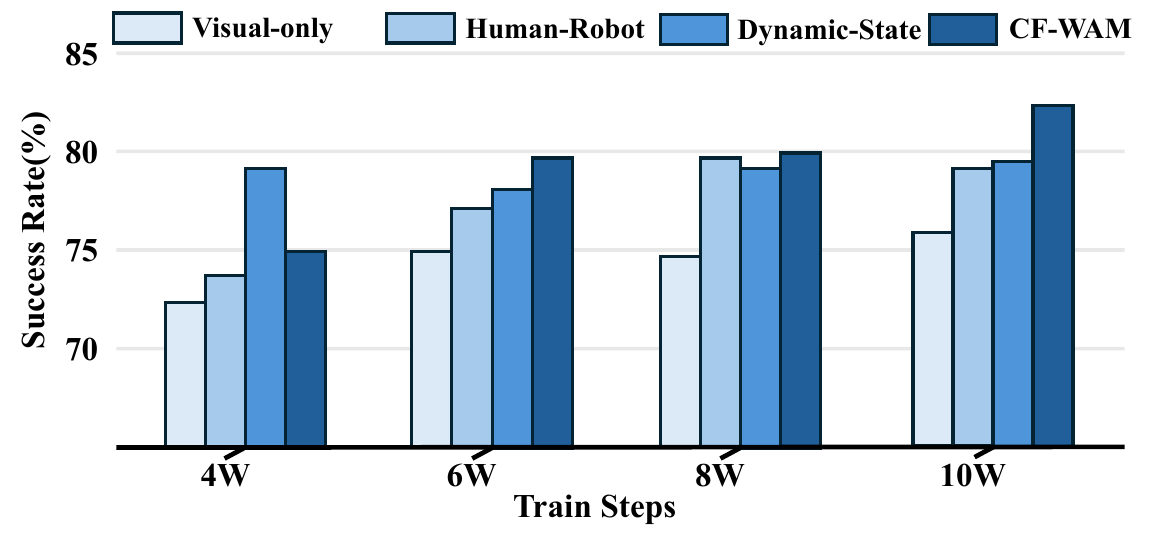}
        \captionof{figure}{
        Success rates at different training steps on RoboCasa-GR1.
        }
        \label{fig:training_efficiency}
    \end{minipage}

\end{figure}

\subsection{Comparison with Existing Methods on RoboCasa-GR1}
We first compare CF-WAM with existing methods on RoboCasa-GR1. As shown in Table~\ref{tab:robocasa_main}, CF-WAM achieves an average success rate of \(82.50\%\) across all 24 tasks, outperforming the strongest prior WAM baseline, WALA, by \(7.33\) percentage points. On novel pick-and-place tasks, CF-WAM reaches \(85.00\%\), improving over WALA by \(11.55\) points and outperforming existing VLA-based methods. These results demonstrate strong overall control performance, with particularly pronounced gains on novel manipulation tasks. Representative closed-loop rollouts are shown in Fig.~\ref{fig:robocasa_rollouts}.

\subsection{Ablation Studies and Analysis}
\subsubsection{Effect of Dynamic Future-State Supervision}
We compare dynamic future-state supervision with fixed single-state supervision under the same model, data, optimization, and training budget. Single-state variants use only Visual, Semantic, Geometric, or Interaction supervision, while Dynamic-State samples multiple future-state projections during training and uses Visual at inference. Each single-state variant uses its corresponding representation for both training and inference. As shown in Table~\ref{tab:robocasa_ablation}, Dynamic-State reaches \(79.00\%\), outperforming the best single-state variant at \(76.25\%\) by \(2.75\) percentage points.

We progressively expand the projection set used for dynamic supervision. Starting from Visual-only at \(76.25\%\), adding Geometric, then Interaction, and finally all four projections improves performance to \(77.50\%\), \(78.33\%\), and \(79.00\%\), respectively. 
This consistent gain is not driven by any single representation, but by progressively enriching the same underlying future-state manifold. Different coordinate systems expose complementary action-relevant constraints from the same action-conditioned future, which accumulate across training steps to shape the WAM.

\subsubsection{Effect of Human Experience}
We further examine the effect of Human experience. Under the same Visual future-state supervision, adding unpaired Human data improves average success from \(76.25\%\) to \(78.25\%\). When combined with dynamic future-state supervision, CF-WAM further reaches \(82.50\%\), gaining another \(4.25\) points. This suggests that the key is not simply more Human data, but how cross-embodiment transitions are represented: dynamic projections expose more transferable semantic, geometric, and interaction regularities than fixed RGB supervision, making Human experience more effective for learning shared world dynamics. Additional action-space ablations are provided in Appendix~A.3.

\subsubsection{Comparison with Static Multi-State Supervision}
We compare Dynamic-State with a Static Multi-State variant that jointly predicts all four future-state representations at each step. Under the same data and compute budget, Static Multi-State achieves only \(68.75\%\), far below Dynamic-State at \(79.00\%\). 
Under this compute-matched setting, simultaneously optimizing multiple prediction spaces introduces stronger optimization interference, causing representation-specific objectives to compete for limited action-conditioning capacity.
 Dynamic-State avoids this competition by supervising one projection at a time while accumulating constraints across updates. Implementation details and attention visualizations are provided in Appendix~A.3.

\begin{table}[t]
\centering

\begin{minipage}[t]{0.50\linewidth}
\centering
\captionof{table}{
Main results on RoboCasa-GR1. Success rates (\%) across 24 tasks.
}
\label{tab:robocasa_main}

\setlength{\tabcolsep}{1.5pt}
\renewcommand{\arraystretch}{1.03}

\begin{tabularx}{\linewidth}{@{}lXccc@{}}
\toprule
Group & Method & Artic. & Novel PnP & Avg. \\
\midrule

\multirow{5}{*}{VLA}
& CoRE-VLA     & 56.20 & 56.60 & 56.50 \\
& ABot-M0     & 61.67 & 57.18 & 58.30 \\
& JoyAI-RA 0.1  & 71.67 & 60.38 & 63.20 \\
& DIAL        & 74.30 & 68.90 & 70.20 \\
& ACE-Ego-0    & 59.00 & 77.40 & 72.80 \\

\cmidrule(lr){1-5}

\multirow{4}{*}{WAM}
& DiT4DiT  & 50.33 & 50.96 & 50.80 \\
& LDA-1B   & 56.33 & 50.09 & 55.40 \\
& FastWAM  & 67.33 & 73.56 & 72.00 \\
& WALA    & \textbf{80.33} & 73.45 & 75.17 \\

\cmidrule(lr){1-5}

\textbf{Ours}
& \textbf{CF-WAM}
& 75.00
& \textbf{85.00}
& \textbf{82.50} \\

\bottomrule
\end{tabularx}
\end{minipage}%
\hspace{0.015\linewidth}%
\begin{minipage}[t]{0.39\linewidth}
\centering
\captionof{table}{
Ablation results on RoboCasa-GR1.
}
\label{tab:robocasa_ablation}

\setlength{\tabcolsep}{1.5pt}
\renewcommand{\arraystretch}{1.03}

\begin{tabularx}{\linewidth}{@{}lXc@{}}
\toprule
Group & Method & Avg. \\
\midrule

\multirow{4}{*}{Single-State}
& Visual      & 76.25 \\
& Semantic    & 73.92 \\
& Geometric   & 73.67 \\
& Interaction & 75.92 \\

\cmidrule(lr){1-3}

\multirow{3}{*}{Dynamic-State}
& Vis.+Geo.        & 77.50 \\
& Vis.+Geo.+Int. & 78.33 \\
& All            & 79.00 \\

\cmidrule(lr){1-3}

Static Multi-State
& All & 68.75 \\

\cmidrule(lr){1-3}

Human--Robot
& Visual-only & 78.25 \\

\cmidrule(lr){1-3}

\textbf{Ours}
& \textbf{CF-WAM}
& \textbf{82.50} \\

\bottomrule
\end{tabularx}
\end{minipage}

\end{table}

\begin{table}[t]
\centering
\caption{
OOD generalization results on LIBERO-Plus.
We report success rates (\%).
}
\label{tab:liberoplus}

\setlength{\tabcolsep}{2.6pt}
\renewcommand{\arraystretch}{1.03}

\begin{tabular}{@{}llcccccccc@{}}
\toprule
Group & Method
& Camera
& Robot
& Language
& Light
& Background
& Noise
& Layout
& Avg. \\
\midrule

\multirow{3}{*}{VLA}
& GWM-VLA & 57.90 & 54.70 & 89.80 & 95.40 & 90.80 & 72.50 & 77.10 & 76.90 \\
& FoMoVLA & 64.00 & 62.20 & 94.00 & 94.10 & \textbf{96.20} & 82.20 & 79.60 & 80.50 \\

\cmidrule(lr){1-10}

\multirow{3}{*}{WAM}
& FastWAM & 16.26 & 44.13 & 66.82 & 79.77 & 52.60 & 38.54 & 61.38 & 51.36 \\

& GaussianWAM & \textbf{79.11} & 56.52 & 92.18 & 89.84 & 66.17 & 86.26 & 70.55 & 77.30 \\

\cmidrule(lr){1-10}

\textbf{Ours}
& \textbf{CF-WAM}
& 66.92
& \textbf{84.90}
& 90.50
& \textbf{98.25}
& 67.84
& \textbf{88.94}
& 81.11
& \textbf{82.65} \\

\bottomrule
\end{tabular}
\end{table}

\begin{figure}[t]
    \centering
    \includegraphics[width=0.9\linewidth]{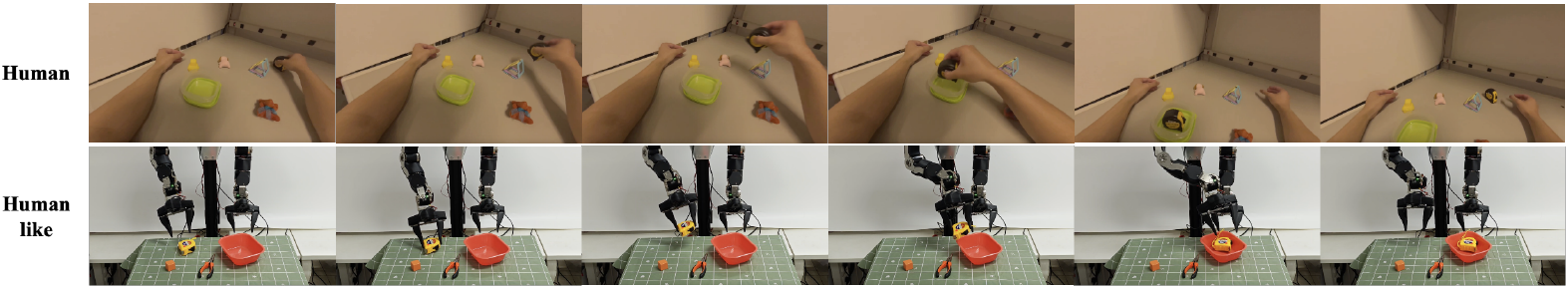}
    \caption{Representative Human and Human-like manipulation trajectories in real-world scenes.}
    \label{fig:realworld}
\end{figure}

\begin{table*}[t!]
\centering
\caption{
Real-world manipulation results.
We report success rates (\%) across six manipulation tasks.
}
\label{tab:realworld_main}

\setlength{\tabcolsep}{4.0pt}
\renewcommand{\arraystretch}{1.05}

\begin{tabular}{@{}lccccccc@{}}
\toprule
Method
& Chemistry
& Folding
& Organization
& Fruit
& Insertion
& Wiping
& Avg. \\
\midrule

$\pi_{0.5}$
& 48.00 & 72.00 & 56.00 & 74.00 & 42.00 & 82.00 & 62.33 \\

GR00T-N1.6
& 32.00 & 44.00 & 38.00 & 72.00 & 34.00 & 64.00 & 47.33 \\

FastWAM
& 16.00 & 62.00 & 40.00 & 46.00 & 22.00 & 64.00 & 41.67 \\

LingBot-VA
& 28.00 & 12.00 & 42.00 & 58.00 & 24.00 & 58.00 & 37.00 \\

\cmidrule(lr){1-8}

CF-WAM (Visual)
& 76.00 & \textbf{92.00} & 74.00 & 88.00 & 76.00 & \textbf{96.00} & 83.67 \\

CF-WAM (Semantic)
& 64.00 & 78.00 & \textbf{76.00} & \textbf{90.00} & 62.00 & 88.00 & 76.33 \\

CF-WAM (Geometric)
& \textbf{88.00} & 86.00 & 70.00 & 86.00 & 74.00 & 88.00 & 82.00 \\

CF-WAM (Interaction)
& 80.00 & 88.00 & 74.00 & 88.00 & \textbf{80.00} & 94.00 & \textbf{84.00} \\

\cmidrule(lr){1-8}

CF-WAM (Oracle)
& 88.00 & 92.00 & 76.00 & 90.00 & 80.00 & 96.00 & 87.00 \\

\bottomrule
\end{tabular}
\end{table*}

\subsubsection{Training Efficiency}
We compare performance across training stages in Fig.~\ref{fig:training_efficiency}. Dynamic-State shows the strongest early performance, suggesting that dynamic future-state supervision accelerates the learning of action-relevant dynamics. CF-WAM initially trails Dynamic-State at \(40\mathrm{K}\) steps but progressively surpasses it as training proceeds, reaching the best final performance at \(100\mathrm{K}\). This suggests that Human experience requires sufficient optimization to fully complement dynamic future-state supervision.

\subsection{OOD Generalization on LIBERO-Plus}
As shown in Table~\ref{tab:liberoplus}, CF-WAM achieves \(82.65\%\) average success on LIBERO-Plus, surpassing FoMoVLA (\(80.50\%\)) and GaussianWAM (\(77.30\%\)). It performs strongly under Robot, Light, and Noise shifts, reaching \(84.90\%\), \(98.25\%\), and \(88.94\%\), indicating robust OOD generalization.

\subsection{Real-World Experiments}

We evaluate CF-WAM on six real-world manipulation tasks covering diverse semantic, geometric, and interaction challenges. As shown in Fig.~\ref{fig:realworld}, we further introduce practical OOD perturbations, including appearance changes, unseen objects, occlusions, and human-like scenes. Detailed settings and OOD results are provided in Appendix~A.3.

As shown in Table~\ref{tab:realworld_main}, all CF-WAM variants substantially outperform existing baselines, with Interaction achieving the best single-projection average of \(84.00\%\). Different projections favor different task characteristics: Geometric performs best on Chemistry, where precise \(3\mathrm{D}\) structure is important; Semantic excels on Organization and Fruit, which rely more on object identity; Interaction performs best on Insertion, where contact evolution is critical; and Visual is strongest on Folding and Wiping, where appearance cues sufficiently describe task progress. Selecting the best projection for each task further improves the average to \(87.00\%\), suggesting the potential of task-adaptive inference.

\section{Conclusion}
We revisit next-state prediction in WAMs and argue that the future-state representation should be dynamic rather than predefined. CF-WAM treats visual, semantic, geometric, and interaction states as complementary projections of the same action-conditioned future, allowing their constraints to accumulate in a unified model while supporting unpaired Human--Robot learning. Experiments across simulation, OOD, and real-world settings demonstrate improved control, generalization, and more effective use of Human experience, highlighting dynamic future-state prediction as a promising direction for world-action learning.

\bibliography{ano_conference}
\bibliographystyle{ano_conference}

\appendix
\section{Appendix}

\subsection{Model Architecture and Implementation Details}
\label{app:model_architecture}

CF-WAM follows a Mixture-of-Transformers (MoT) architecture comprising a video expert and an action expert. The video expert is initialized from Wan2.2-TI2V-5B and models action-conditioned future states, whereas the smaller ActionDiT predicts continuous control trajectories from the current observation, language instruction, and predictive video representations. Both experts contain 30 Transformer layers and perform one asymmetric mixed self-attention operation at each layer.

The experts share the attention computation but not the model parameters. Each expert maintains its own residual stream, attention projections, positional encoding, time conditioning, text cross-attention, feed-forward network, and adaptive-normalization parameters. This parameterization preserves the spatiotemporal generation capability of the pretrained video backbone while allowing the action branch to learn control in a more compact hidden space.

\subsubsection{Expert Configuration}
\label{app:expert_configuration}

The configurations of the video and action experts are summarized in Table~\ref{tab:expert_configuration}. The video expert has a residual width of 3072, with 24 attention heads of dimension 128 and an FFN intermediate dimension of 14336. Video tokens use three-dimensional rotary positional embeddings (RoPE) over temporal and spatial coordinates $(t,h,w)$. The video expert contains approximately $5.000$B parameters.

The action expert has a residual width of 1024 and an FFN intermediate dimension of 4096, amounting to approximately $1.021$B parameters. Its input is a 16-step action chunk with 47 dimensions per step. Each action step is represented by one token, and one-dimensional RoPE encodes its relative position within the prediction horizon.

Although the experts use asymmetric residual widths, both employ a 3072-dimensional representation inside self-attention, organized as 24 heads of dimension 128. They can therefore interact in a common multi-head attention space while retaining separate residual streams for cross-attention and feed-forward computation. Excluding the frozen VAE and text encoder, the two Transformer experts contain approximately $6.021$B parameters in total.

\begin{table}[t]
    \centering
    \caption{Expert configurations in CF-WAM. The attention width denotes the feature dimension used for cross-expert mixed attention.}
    \label{tab:expert_configuration}

    \setlength{\tabcolsep}{6pt}

    \begin{tabular}{lcc}
        \toprule
        Configuration & Video Expert & Action Expert \\
        \midrule
        Backbone & Wan2.2-TI2V-5B DiT & ActionDiT \\
        Transformer layers & 30 & 30 \\
        Residual width & 3072 & 1024 \\
        Attention width & 3072 & 3072 \\
        Attention heads & $24\times128$ & $24\times128$ \\
        Positional encoding & 3D RoPE $(t,h,w)$ & 1D RoPE \\
        FFN intermediate width & 14336 & 4096 \\
        Input & Video latent tokens & $16\times47$ action chunk \\
        Output & Video latent velocity & $16\times47$ action velocity \\
        Parameters & $5.000$B & $1.021$B \\
        \bottomrule
    \end{tabular}
\end{table}

\subsubsection{Cross-Expert Mixed Attention}
\label{app:mixed_attention_impl}

Mixed attention aligns the experts in attention space while keeping their parameters independent. Each expert first computes Query, Key, and Value using its own projection layers. The video residual stream already has the same width as the 3072-dimensional attention space. In contrast, the action expert uses its independently parameterized $W_q^{\mathrm a}$, $W_k^{\mathrm a}$, and $W_v^{\mathrm a}$ to project its 1024-dimensional hidden states into the common attention space:
\begin{equation}
    1024 \xrightarrow{\,W_q^{\mathrm a},W_k^{\mathrm a},W_v^{\mathrm a}\,}
    24\times128=3072.
\end{equation}

Before concatenating the two sequences, the Query and Key of each expert are processed by expert-specific RMSNorm layers. The video expert then applies three-dimensional RoPE, whereas the action expert applies one-dimensional RoPE. Positional encoding is therefore completed independently before mixed attention, preserving modality-specific positional structure while mapping both token streams into an attention space with aligned head count and head dimension.

The resulting Query, Key, and Value tensors are concatenated only along the sequence dimension and processed by a single FlashAttention operation. The $h$-th head of the video expert and the $h$-th head of the action expert occupy the corresponding 128-dimensional head space, so no additional feature adapter is required for cross-expert attention. The output is subsequently split according to the original video and action sequence boundaries and mapped back to the corresponding residual spaces using expert-specific output projections:
\begin{equation}
    W_o^{\mathrm v}:\mathbb{R}^{3072}\rightarrow\mathbb{R}^{3072},
    \qquad
    W_o^{\mathrm a}:\mathbb{R}^{3072}\rightarrow\mathbb{R}^{1024}.
\end{equation}

Each projected output is modulated by the adaptive-normalization gate of its expert and added to the corresponding residual stream. The two token streams then proceed through separate text cross-attention and FFN modules. In particular, all subsequent computation in the action branch remains within its 1024-dimensional residual space. CF-WAM thus shares the attention computation rather than the Query, Key, Value, output-projection, or FFN parameters.

We use an asymmetric attention topology. Video queries attend only to video keys and values, whereas action queries attend to all video and action keys and values. Action tokens are mutually visible, while the video stream preserves the causal constraint associated with first-frame conditioning. Consequently, the action expert can read the current observation and predictive state representations encoded by the video expert, while noisy action tokens do not perturb video generation. Gradient checkpointing is enabled for all mixed-attention modules to reduce the memory footprint of training the large video backbone with the joint token sequence.

\subsubsection{Temporal Window and Sequence Construction}
\label{app:tokenization}

The temporal construction and per-sample token budget are summarized in Table~\ref{tab:token_budget}. Each training sample spans a $0.8\,\mathrm{s}$ temporal window. Observations and actions are sampled at 20 Hz, yielding 17 consecutive observation frames and the corresponding 16 control steps between adjacent frames.

The video branch selects five keyframes with indices $\{0,4,8,12,16\}$. Each frame is resized to $224\times224$. The frozen Wan2.2 VAE encodes the clip into a latent tensor with 48 channels, two temporal positions, and a spatial resolution of $28\times28$. We then patchify the latent using a $(1,2,2)$ spatiotemporal patch size, resulting in
\begin{equation}
    N_{\mathrm v}=2\times14\times14=392
\end{equation}
video tokens.

The action branch uses the corresponding 16-step action chunk, with one token for each control step. The total sequence length processed by mixed self-attention is therefore
\begin{equation}
    N_{\mathrm{joint}}=N_{\mathrm v}+N_{\mathrm a}=392+16=408.
\end{equation}
Language and proprioceptive conditions are injected through cross-attention and are not included in this joint self-attention length.

\begin{table}[t]
    \centering
    \caption{Per-sample tokenization and sequence budget.}
    \label{tab:token_budget}
    \setlength{\tabcolsep}{5pt}
    \begin{tabular}{lcc}
        \toprule
        Stream & Input and processing & Tokens \\
        \midrule
        Video input & 5 frames, $224\times224\times3$ & -- \\
        VAE latent & 48 channels, $2\times28\times28$ & -- \\
        Video tokens & Patchify $(1,2,2)$ & 392 \\
        Action chunk & 16 steps, 47 dimensions per step & 16 \\
        Joint self-attention & Video + action tokens & 408 \\
        Language context & umT5-XXL, up to 128 tokens & 128 \\
        Proprioceptive context & 47-dimensional initial state & 1 \\
        \bottomrule
    \end{tabular}
\end{table}

\subsubsection{Conditioning Inputs}
\label{app:conditioning}

Language instructions are encoded by the frozen umT5-XXL text encoder. Each instruction is represented by at most 128 token embeddings with a hidden dimension of 4096. The embeddings are precomputed and remain fixed throughout training.

To identify the requested future-state projection, we append a representation-specific condition to the original task instruction. Depth, interaction-segmentation, and semantic targets respectively use \texttt{Predict the depth video.}, \texttt{Predict the segmentation video.}, and \texttt{Predict the semantic video.}. These conditions are processed together with the task instruction by the existing text cross-attention modules of both experts, allowing a single generative model to switch its future-state output without modality-specific prediction heads.

The 47-dimensional proprioceptive state at the beginning of the trajectory is projected to 4096 dimensions and appended as one additional context token. The cross-attention context therefore contains up to 128 language tokens and one proprioceptive token. For EgoDex samples, unavailable robot joint states are zero-padded to preserve a unified tensor shape, but the corresponding action dimensions are excluded from the training loss.

All future-state projections are conditioned on the time-aligned current RGB observation. The current image is encoded by the frozen Wan2.2 VAE and written into the first temporal position of the video latent. This latent is reinserted at every denoising step to prevent corruption of the conditioning frame by flow-matching noise. Depth, interaction-segmentation, and DINOv3-PCA videos therefore share the same current visual state as their initial condition.

\subsubsection{Flow Matching, Optimization, and Inference}
\label{app:training_details}

The video and action branches maintain independent noisy states, time conditions, and velocity heads. Both branches use the shifted time schedule
\begin{equation}
    \sigma(u)=\frac{5u}{1+4u},
    \qquad u\sim\mathcal{U}(0,1).
\end{equation}
Training time is discretized into 1000 positions, with loss weighting
\begin{equation}
    w(t)\propto
    \exp\!\left[-2\left(\frac{t-500}{1000}\right)^2\right].
\end{equation}

The video loss is computed in the VAE latent space, with temporal and spatial padding masks excluding invalid positions. The action loss is a per-step, per-dimension mean squared error. Teleop-GR1 samples supervise all 47 action dimensions, whereas EgoDex samples supervise only the valid 18-dimensional wrist end-effector actions. The remaining dimensions are zero-padded but masked from the loss. Both the video and action loss weights are set to 1.0.

The training and inference configurations are summarized in Table~\ref{tab:training_configuration}. We optimize CF-WAM using AdamW with an initial learning rate of $1\times10^{-4}$, weight decay of $10^{-2}$, and cosine learning-rate decay. Training uses bf16 precision and ZeRO-1 distributed optimization. The RoboCasa-GR1 model is trained for 100K optimization steps with a global batch size of 32 on four nodes, each equipped with eight NVIDIA B200 GPUs. The Wan2.2 VAE and umT5-XXL text encoder remain frozen. EgoDex and Teleop-GR1 samples are drawn with a sampling ratio of $1{:}2$. We use a uniform projection-sampling distribution \(p_{\mathcal R}(k)=1/4\) for Visual, Semantic, Geometric, and Interaction.

At inference time, the video and action branches are integrated using a 20-step Euler solver with shift $=5$. They share the same integration grid but update their noisy states independently. At each step, the current-RGB latent is reinserted into the first temporal position before cross-expert mixed attention and velocity prediction. The action expert finally produces a 16-step trajectory with 47 action dimensions per step.

\begin{table}[t]
    \centering
    \caption{Training and inference configuration of CF-WAM on RoboCasa-GR1.}
    \label{tab:training_configuration}
    \setlength{\tabcolsep}{6pt}
    \begin{tabular}{ll}
        \toprule
        Item & Configuration \\
        \midrule
        Optimizer & AdamW \\
        Initial learning rate & $1\times10^{-4}$ \\
        Learning-rate schedule & Cosine decay \\
        Weight decay & $10^{-2}$ \\
        Numerical precision & bf16 \\
        Distributed optimization & ZeRO-1 \\
        Compute & 4 nodes $\times$ 8 NVIDIA B200 GPUs \\
        Global batch size & 32 \\
        Training steps & 100K \\
        Flow-time positions & 1000 \\
        Flow shift & 5 \\
        Inference solver & 20-step Euler \\
        Video/action loss weights & $1.0/1.0$ \\
        Frozen components & Wan2.2 VAE, umT5-XXL \\
        Data-source sampling ratio & EgoDex : Teleop-GR1 $=1{:}2$ \\
        Projection sampling & Vis. : Sem. : Geo. : Int. $=1{:}1{:}1{:}1$ \\
        \bottomrule
    \end{tabular}
\end{table}

\subsection{Future-State Projection Data Construction}
\label{app:data_construction}

CF-WAM is trained on simulated Robot teleoperation trajectories from Teleop-GR1 and real egocentric Human manipulation trajectories from EgoDex. For each physical trajectory, we construct four aligned future-state projections: Visual (RGB), Semantic (DINOv3-PCA), Geometric (depth), and Interaction (segmentation). This produces eight source--projection subsets:
\begin{equation}
    \{\mathrm{EgoDex},\mathrm{Teleop\text{-}GR1}\}
    \times
    \{\mathrm{RGB},\mathrm{DINOv3\text{-}PCA},\mathrm{Depth},\mathrm{Seg}\}.
\end{equation}

As summarized in Table~\ref{tab:multimodal_data}, after quality filtering, EgoDex contains 23,849 aligned episodes, while Teleop-GR1 contains 24 RoboCasa-GR1 tabletop tasks with 1,000 episodes per task, totaling 24,000 episodes. Both sources are standardized to 20 Hz and represented with a common 47-dimensional action tensor.

The four projections are alternative descriptions of the same physical trajectory rather than independent samples. Within each source, corresponding episode, frame, and action indices refer to the same physical instant. For non-RGB projections, the strictly time-aligned current RGB observation is used as the first-frame condition, while subsequent frames are represented in the sampled future-state projection. Thus, all projections share the same current world state and differ only in how the future is represented.

\begin{table}[t]
    \centering
    \caption{Composition of the core multimodal training mixture. The number of modality episodes counts the four future-state representations associated with each physical trajectory.}
    \label{tab:multimodal_data}
    \small
    \setlength{\tabcolsep}{4pt}
    \resizebox{\linewidth}{!}{%
    \begin{tabular}{lcccccc}
        \toprule
        Source & Data type & Physical episodes & Modality episodes & Representations & FPS & Action \\
        \midrule
        EgoDex & Real egocentric human & 23,849 & 95,396 & RGB/Depth/Seg/DINOv3-PCA & 20 & 18 valid, padded to 47 \\
        Teleop-GR1 & Simulated robot teleoperation & 24,000 & 96,000 & RGB/Depth/Seg/DINOv3-PCA & 20 & 47 dimensions \\
        \midrule
        Total & -- & 47,849 & 191,396 & 8 modality subsets & 20 & Unified to 47 \\
        \bottomrule
    \end{tabular}}
\end{table}

\subsubsection{Simulated Teleop-GR1 Data}
\label{app:teleop_data}

\paragraph{Data source.}
The Robot data are obtained from the RoboCasa-GR1 tabletop portion of NVIDIA's PhysicalAI-Robotics-GR00T-Teleop-Sim-XEmbodiment-Compatible dataset. The dataset contains teleoperated Fourier GR-1 trajectories in RoboCasa and covers 24 tabletop tasks, with 1,000 episodes per task.

The original trajectories provide 44-dimensional Robot joint states and actions but do not directly contain end-effector trajectories that can be expressed in the same representation as Human wrist motion. We therefore replay the recorded Robot states in simulation and recover the poses of both wrist end effectors at every time step, providing a common end-effector representation for Human--Robot learning.

\paragraph{End-effector trajectory recovery.}
For each episode, the Robot is replayed frame by frame according to the recorded joint states. We obtain the pose of each wrist frame relative to the robot base frame. Each wrist is represented by a three-dimensional position and a continuous six-dimensional rotation representation:
\begin{equation}
    a_{\mathrm{wrist}}
    = [p_x,p_y,p_z,r_1,\ldots,r_6]
    \in\mathbb{R}^{9}.
    \label{eq:wrist_representation}
\end{equation}
The two wrists jointly form an 18-dimensional end-effector representation:
\begin{equation}
    a_{\mathrm{EE}}
    = [a_{\mathrm{wrist}}^{\mathrm r},a_{\mathrm{wrist}}^{\mathrm l}]
    \in\mathbb{R}^{18}.
    \label{eq:ee_representation}
\end{equation}
The continuous six-dimensional rotation representation avoids the discontinuities of Euler angles and provides a continuous regression space. The recovered trajectories are aligned with the original joint actions by episode and time step.

\paragraph{Unified action space.}
The final Teleop-GR1 action vector is ordered as
\begin{equation}
    a = [
    a_{\mathrm{wrist}}^{\mathrm r},
    a_{\mathrm{wrist}}^{\mathrm l},
    a_{\mathrm{r\_arm}},
    a_{\mathrm{r\_hand}},
    a_{\mathrm{l\_arm}},
    a_{\mathrm{l\_hand}},
    a_{\mathrm{waist}}],
\label{eq:teleop_action}
\end{equation}
with dimensionality
\begin{equation}
    47 = 9+9+7+6+7+6+3.
\label{eq:teleop_action_dim}
\end{equation}
The first 18 dimensions encode the two wrist poses. The remaining 29 dimensions represent the two arms, two dexterous hands, and waist joints. These Robot-specific dimensions are selected from the original 44-dimensional action and reordered according to the unified control convention. After conversion, we recompute state and action statistics and verify alignment among end-effector trajectories, joint actions, RGB observations, and timestamps.

\paragraph{Future-state projection construction and alignment.}
For Teleop-GR1, RGB, depth, segmentation, and DINOv3-PCA videos form four aligned future-state projections of the same physical trajectory. All projections are associated with the converted 47-dimensional Robot trajectories using the same episode and temporal indices. For non-RGB projections, the strictly time-aligned current RGB observation is used as the first-frame condition, while subsequent frames are represented in the corresponding future-state projection. Thus, all projections share the same current world state and differ only in how the future is represented.

\paragraph{Simulation segmentation.}
The segmentation projection is generated directly from MuJoCo rather than predicted by an external perception model. For each Teleop-GR1 episode, we first locate the corresponding original GR00T demonstration and recover its MuJoCo model together with the recorded simulator states. Each video frame is associated with a simulator state according to
\begin{equation}
    i_s
    =
    \operatorname{round}
    \left(
    \frac{i_f}{N_f-1}(N_s-1)
    \right),
\label{eq:sim_frame_state_mapping}
\end{equation}
where $i_f$ and $i_s$ denote the video-frame and simulator-state indices, and $N_f$ and $N_s$ are their respective sequence lengths. We restore the corresponding $qpos$ and $qvel$, execute a MuJoCo forward pass, and render the egocentric camera in segmentation mode. This produces a per-pixel geometry ID map, which is subsequently mapped to task-relevant interaction roles according to the associated geometry and body identities.

We retain the target object, manipulated object, left manipulator, and right manipulator as the interaction categories. The left hand and left arm are merged into the left-manipulator role, while the right hand and right arm are merged analogously. All remaining pixels are treated as background. The resulting masks are encoded using a fixed role palette:
\begin{equation}
\begin{aligned}
\mathrm{background} &:\ (0,0,0),\\
\mathrm{target} &:\ (0,0,255),\\
\mathrm{manipulated} &:\ (255,0,0),\\
\mathrm{left} &:\ (0,255,0),\\
\mathrm{right} &:\ (255,255,0).
\end{aligned}
\label{eq:sim_seg_palette}
\end{equation}
Thus, the segmentation videos contain pure role labels rather than RGB overlays. The masks are stored frame by frame and rendered into lossless RGB videos so that the categorical colors are preserved exactly. The simulator pipeline therefore provides interaction-state supervision from geometry-level ground truth rather than from estimated visual masks.

\paragraph{Simulation depth.}
The geometric projection is generated through the same simulator-state replay procedure. Instead of segmentation rendering, we enable MuJoCo depth rendering for the egocentric camera, obtaining metric camera-$Z$ depth in meters. To maintain pixel-level correspondence with the RGB observations, the rendered depth maps follow the same camera preprocessing used by the co-training visual stream:
\begin{equation}
1280\times800
\rightarrow
\mathrm{crop}(310{:}770,\ 110{:}1130)
\rightarrow
720\times480
\rightarrow
720\times720
\rightarrow
256\times256.
\label{eq:sim_camera_transform}
\end{equation}
The intermediate $720\times720$ image is obtained by padding the $720\times480$ crop vertically. Nearest-neighbor interpolation is used for depth resizing to avoid introducing interpolated values across object boundaries. The resulting metric depth is quantized in millimeters and stored as a 16-bit depth map together with a validity mask.

\paragraph{Common depth-video encoding.}
To express metric depth using the same three-channel video interface as the other future-state projections, we convert the stored millimeter depth values to meters and apply a nonlinear mapping
\begin{equation}
    d = \frac{d_{\mathrm{mm}}}{1000},
    \qquad
    u
    =
    1-
    \left(
    1+\frac{d}{3}
    \right)^{-2},
\label{eq:wam_depth_mapping}
\end{equation}
where $u\in[0,1)$ denotes the normalized depth coordinate. The $3\,\mathrm{m}$ parameter controls the scale of the nonlinear mapping rather than defining a maximum depth, so depths beyond $3\,\mathrm{m}$ are not truncated. This transformation allocates finer numerical resolution to nearby regions while progressively compressing farther distances.

The normalized coordinate is then mapped continuously along a fixed RGB path,
\begin{equation}
\begin{aligned}
&(0,0,0)
\rightarrow
(0,0,255)
\rightarrow
(0,255,255)
\rightarrow
(0,255,0)\\
&\qquad\rightarrow
(255,255,0)
\rightarrow
(255,0,0)
\rightarrow
(255,0,255)
\rightarrow
(255,255,255),
\end{aligned}
\label{eq:wam_depth_palette}
\end{equation}
corresponding to black, blue, cyan, green, yellow, red, magenta, and white. Invalid depth pixels are encoded as black. The resulting videos are stored using lossless RGB encoding, preventing compression-induced changes to the depth colors. This same depth-video parameterization is used across data sources, so Geometric supervision has a consistent video representation despite different depth-generation procedures.

\subsubsection{Egocentric Human Data from EgoDex}
\label{app:egodex_data}

EgoDex consists of real egocentric Human manipulation trajectories. Its original state and action representation is 621-dimensional, covering wrist, hand, and other Human-body states. CF-WAM extracts the two wrist motions that can be mapped to the Robot end-effector representation and discards Human-body variables without a direct counterpart in Robot control.

\paragraph{Wrist-action extraction.}
At each time step, the nine-dimensional right- and left-wrist poses are extracted as
\begin{equation}
    a_{\mathrm{right}}=a^{\mathrm{raw}}[324{:}333],
    \qquad
    a_{\mathrm{left}}=a^{\mathrm{raw}}[36{:}45].
\label{eq:egodex_wrist_extraction}
\end{equation}
Each wrist comprises a three-dimensional position and a six-dimensional rotation representation. The resulting Human action is
\begin{equation}
    a_{\mathrm{human}}
    =
    [a_{\mathrm{right}},a_{\mathrm{left}}]
    \in\mathbb{R}^{18}.
\label{eq:egodex_human_action}
\end{equation}

The extracted wrist poses are mapped to the Robot end-effector convention through fixed ARKit-to-GR1 coordinate transformations. The transformation consists of a global coordinate mapping together with hand-specific local transformations, applied consistently to both position and rotation. The transformed 18-dimensional wrist action occupies the first 18 dimensions of the unified 47-dimensional action tensor. The remaining 29 Robot-specific joint dimensions are zero-padded and excluded from the action loss through a source-dependent mask.

\paragraph{Temporal synchronization and spatial standardization.}
The original EgoDex videos and actions are recorded at 30 Hz. To convert them to the 20 Hz control rate, the source-frame index associated with resampled frame $j$ is
\begin{equation}
    i_j=\operatorname{round}(1.5j).
\label{eq:egodex_temporal_resampling}
\end{equation}
Video, action, and state sequences use the same index set, avoiding visual--action misalignment caused by independent resampling. Frame indices, timestamps, and episode lengths are reconstructed after resampling. The processed videos are stored at a constant 20 Hz with a spatial resolution of $456\times256$ and are further resized to $224\times224$ when loaded for WAM training.

\paragraph{EgoDex segmentation construction.}
Since EgoDex does not provide simulator-ground-truth interaction masks, we construct the Interaction projection using a perception-based annotation pipeline. Given the task instruction, Qwen is first used to identify task-relevant entities and localize them on sparse keyframes sampled at approximately 25\% and 75\% of the video. For task objects, Qwen provides bounding-box prompts. For the left and right hands, it additionally provides interior hand points together with bare-arm and sleeve points, while points from the opposite hand are used as negative prompts to reduce cross-hand mask leakage.

The resulting prompts are passed to SAM3, which initializes object tracks on the keyframes and propagates the masks bidirectionally over the full video. The raw annotations are subsequently converted into a compact set of interaction roles used by CF-WAM: manipulated object, target object, left hand, and right hand. All remaining pixels are assigned to the background. The four interaction roles use the same fixed color convention as the simulation data, so Human and Robot trajectories share a consistent Interaction-state representation.

\paragraph{Segmentation quality control and cross-projection alignment.}
The propagated masks are automatically checked for common failure cases, including missing target objects, abrupt mask discontinuities, inconsistent appearance, abnormally small manipulated-object motion, and excessively large hand masks. When such failures are detected, the annotation pipeline performs an automatic repair pass using renewed Qwen localization and SAM3 propagation. Episodes that remain invalid after repair are discarded.

Starting from 27,419 EgoDex episodes, 3,570 episodes are removed by this quality-control procedure, leaving 23,849 valid trajectories. We use this valid subset as the common episode index for RGB, depth, segmentation, and DINOv3-PCA projections. Consequently, the four projections share the same episode set, trajectory order, frame count, action indices, task labels, and sample-level temporal correspondence.

\paragraph{EgoDex depth construction.}
The Geometric projection for EgoDex is constructed offline using Depth Anything 3 (DA3). Each RGB frame is processed by DA3 at an inference resolution of 504 pixels, with metric output enabled and ray-pose conditioning disabled. The predicted depth is rendered at $256\times256$ resolution.

To convert the DA3 output into metric depth, we use the camera intrinsics stored in each EgoDex HDF5 episode. The intrinsic matrix is validated before use, including correction of transposed matrices in legacy files. The calibrated EgoDex camera has
\begin{equation}
    f_x=f_y=736.633911,
    \qquad
    c_x=960,
    \qquad
    c_y=540,
\label{eq:egodex_camera_intrinsics}
\end{equation}
at the original $1920\times1080$ image resolution. The focal length is rescaled to the depth-map resolution as
\begin{equation}
    \tilde f_x
    =
    f_x\frac{W_d}{1920},
    \qquad
    \tilde f_y
    =
    f_y\frac{H_d}{1080},
    \qquad
    \tilde f
    =
    \frac{\tilde f_x+\tilde f_y}{2},
\label{eq:egodex_scaled_focal}
\end{equation}
where $W_d$ and $H_d$ denote the depth-map width and height. The raw DA3 prediction is then converted to metric depth by
\begin{equation}
    d
    =
    d_{\mathrm{DA3}}
    \frac{\tilde f}{300},
\label{eq:egodex_metric_depth}
\end{equation}
where $d$ is measured in meters. The resulting metric depth is quantized in millimeters and stored as a 16-bit depth map together with a validity mask.

The depth sequence initially follows the original 30 Hz EgoDex frame timing. It is resampled to 20 Hz using exactly the same frame indices as the RGB, action, and state sequences. The $256\times256$ depth maps are then resized to $456\times256$ using nearest-neighbor interpolation to match the processed EgoDex video geometry without introducing invalid intermediate depth colors. During WAM training, they are further resized to the common $224\times224$ model input resolution.

The metric depth is subsequently converted into the common three-channel depth-video representation defined in Eq.~\ref{eq:wam_depth_mapping} and Eq.~\ref{eq:wam_depth_palette}. Thus, although EgoDex depth is estimated by DA3 whereas Teleop-GR1 depth is rendered directly from MuJoCo, both sources use the same Geometric future-state parameterization for WAM supervision.

\subsubsection{Real-Robot Data}
\label{app:real_robot_data}

The real-world experiments additionally use real-Robot manipulation trajectories collected on the physical platform. Each trajectory contains synchronized RGB observations, Robot actions, and task instructions. The Robot-side action representation follows the control convention used in our real-world policy training, while all visual observations are converted into the same four future-state projections used for EgoDex and Teleop-GR1.

\paragraph{Future-state projection construction.}
The Real-Robot data follow the same projection-construction pipeline as EgoDex. RGB observations are used directly as the Visual projection. The Interaction projection is generated using the same Qwen-assisted localization and SAM3 mask-propagation pipeline, with task-relevant objects and Robot manipulators mapped to the corresponding interaction roles. The Geometric projection is estimated using Depth Anything 3 with the camera intrinsics of the real-Robot system and is converted into the common three-channel depth-video representation defined in Eq.~\ref{eq:wam_depth_mapping} and Eq.~\ref{eq:wam_depth_palette}. The Semantic projection is generated using the same frozen DINOv3 teacher and fixed global PCA mapping described in Sec.~\ref{app:dinov3_pca}.

\paragraph{Temporal and cross-projection alignment.}
RGB observations, Robot actions, and all constructed future-state projections are synchronized using the same trajectory timestamps. The four projections are standardized using the same temporal and spatial preprocessing as the EgoDex data and retain identical episode and frame correspondence. For non-RGB projections, the time-aligned current RGB observation is used as the first-frame condition, ensuring that all projections start from the same current world state and differ only in their representation of the future.

\subsubsection{Global DINOv3-PCA Semantic Videos}
\label{app:dinov3_pca}

CF-WAM represents semantic future states using three-channel DINOv3-PCA videos. These videos are generated offline using a frozen DINOv3 teacher and are processed by the same frozen VAE as the other future-state projections. Instead of fitting PCA independently for each image or frame, we fit a single fixed global projection over the full dataset. This avoids temporal inconsistency caused by PCA component permutation and sign ambiguity and establishes a consistent semantic coordinate system across data sources and time.

\paragraph{Fitting the global PCA mapping.}
The frozen DINOv3 teacher processes each frame at the configured 1280-pixel input resolution and extracts patch tokens in BF16. To reduce the influence of large static background regions on the global PCA basis, a foreground-probability predictor is used to identify candidate foreground tokens with a threshold of 0.5, followed by a $3\times3$ median filter to remove isolated noise. The PCA fitting set is constructed from foreground DINOv3 features extracted from all frames of all available EgoDex and Teleop-GR1 videos, without episode- or frame-level subsampling.

We use randomized PCA with three whitened components:
\begin{equation}
    \hat f=(f-\mu)C\Lambda^{-1/2},
\label{eq:global_dino_pca}
\end{equation}
where $\mu$ is the global feature mean, $C$ contains the first three PCA components, and $\Lambda^{-1/2}$ contains the corresponding whitening scales. The random seed is fixed to 42. To remove the sign ambiguity of PCA, each component is oriented such that its element with the largest absolute magnitude is positive.

The three whitened components are mapped to RGB values as
\begin{equation}
    y^{\mathrm{Sem}}
    =
    \operatorname{sigmoid}(2.0\,\hat f).
\label{eq:dino_semantic_rgb}
\end{equation}
The fitted global mean, PCA components, and whitening scales remain fixed for all subsequent semantic-video rendering.

\paragraph{Semantic-video rendering.}
Semantic videos are precomputed using multi-GPU distributed inference, with one frozen DINOv3 teacher instantiated on each GPU. For each frame, DINOv3 patch tokens are extracted at the 1280-pixel teacher resolution, transformed using the fixed global PCA mapping, and converted to three-channel values using $\operatorname{sigmoid}(2.0\,\cdot)$. The resulting patch representation forms an $80\times80$ RGB grid, which is resized to $224\times224$ and encoded as an episode-level video.

Precomputation covers all 27,419 EgoDex episodes and all 24,000 Teleop-GR1 episodes. The EgoDex semantic videos are subsequently filtered to the same 23,849 episodes retained by the segmentation quality-control stage, ensuring a common trajectory set across all four future-state projections.

\paragraph{Training interface.}
DINOv3-PCA representations are stored as three-channel $224\times224$ videos and follow the same VAE encoding path as the other future-state projections. The DINOv3 teacher is used only for offline semantic-target construction and does not participate in CF-WAM training or inference. Each semantic sequence uses the strictly time-aligned current RGB observation as its first-frame condition, followed by DINOv3-PCA future frames, and is conditioned on the language suffix \emph{Predict the semantic video.}

\subsubsection{Cross-Projection Consistency Checks}
\label{app:projection_alignment}

Before mixed training, we perform a common set of consistency checks across all future-state projections. For each data source and episode, we verify that: (1) all projections contain the same episodes; (2) video frame counts agree with trajectory lengths; (3) video lengths match across projections; (4) observations, actions, and states use the same temporal indices; (5) corresponding sample indices refer to the same physical instant; (6) every non-RGB sequence starts from the strictly time-aligned current RGB observation; (7) actions and states are mapped to the unified dimensions; and (8) padded invalid action dimensions are correctly excluded by the loss mask.

These checks ensure consistency at three levels: temporal alignment among observations, actions, and states; physical correspondence across future-state projections; and a common Human--Robot action tensor with source-dependent masks indicating valid control dimensions.

\subsection{Additional Experimental Details and Results}
\label{app:additional_experiments}

\subsubsection{Static Multi-State Baseline Implementation}
\label{app:static_multistate}
\paragraph{Static Multi-State baseline implementation.}
To isolate the effect of dynamic future-state parameterization, we construct a Static Multi-State baseline that predicts all four future-state projections simultaneously for every training sample. The baseline uses the same Visual, Semantic, Geometric, and Interaction targets as CF-WAM, corresponding to RGB, DINOv3-PCA, depth, and segmentation videos, respectively. Unlike Dynamic-State, which samples one projection for each training sample, Static Multi-State explicitly maintains all four prediction spaces within every update.

\paragraph{Four-projection sample construction.}
For each physical trajectory, the dataset retrieves four strictly paired future videos from the RGB, depth, segmentation, and DINOv3-PCA subsets using the same source, episode, and temporal index. The four videos are required to have identical tensor shapes, frame-padding masks, and task instructions; any inconsistency is treated as a data error rather than silently remapped to another sample. A single action trajectory, proprioceptive state, and task instruction are taken from the corresponding RGB sample and shared across the four future-state projections. Thus, each training example contains four alternative descriptions of exactly the same future transition together with only one action target.

To ensure that the four prediction branches differ only in their representation of the future, the first frame of every non-RGB sequence is replaced by the strictly time-aligned current RGB observation. Consequently, all four streams are conditioned on the same current world state, while their subsequent frames correspond to different future-state coordinate systems. This is the same first-frame conditioning convention used by CF-WAM. 

\paragraph{Model architecture.}
Static Multi-State reuses the same video expert, action expert, VAE, text-conditioning pathway, and action representation as CF-WAM. Importantly, the baseline introduces no additional learnable parameters: the four-projection implementation only changes how the existing video and action streams are organized during attention, while the underlying Transformer parameters remain unchanged. The resulting model therefore has the same learnable parameterization as the corresponding single-projection WAM and can be initialized from the same checkpoint. 

For a batch of size $B$, the four future videos are encoded by the same frozen VAE and organized as four latent streams. Each stream is processed by the same video expert, while maintaining its own latent state. At every Transformer layer, each video query attends only to the keys and values from its own projection. In particular, Visual, Semantic, Geometric, and Interaction video tokens do not directly attend to one another. This prevents the baseline from introducing an additional cross-projection video-fusion module.

The action branch remains a single stream. Its queries attend jointly to the concatenated key/value tokens from all four video projections together with the action stream itself. Therefore, the action expert can simultaneously access information exposed by all four future-state representations, while the four video streams remain independently modeled. The video streams do not attend to action tokens, preserving the asymmetric world-to-action attention topology used by CF-WAM. 

\paragraph{Flow matching and training objective.}
The four video projections share the same sampled video flow time at each update so that they are supervised at the same noise level, while independent noise realizations are sampled for their latent states. The action branch maintains its own flow time and noise state, following the same world-action flow-matching formulation as CF-WAM. The current-RGB latent is kept clean and clamped to the first temporal position of all four video streams throughout training.

A separate video prediction loss is computed for each future-state projection. Let
$\mathcal{L}_{\mathrm{Vis}}$,
$\mathcal{L}_{\mathrm{Sem}}$,
$\mathcal{L}_{\mathrm{Geo}}$, and
$\mathcal{L}_{\mathrm{Int}}$
denote the four video losses. We assign equal weight to all projections and optimize
\begin{equation}
    \mathcal{L}_{\mathrm{static}}
    =
    \lambda_w
    \frac{1}{4}
    \left(
    \mathcal{L}_{\mathrm{Vis}}
    +
    \mathcal{L}_{\mathrm{Sem}}
    +
    \mathcal{L}_{\mathrm{Geo}}
    +
    \mathcal{L}_{\mathrm{Int}}
    \right)
    +
    \lambda_a
    \mathcal{L}_{\mathrm{action}}.
\label{eq:static_multistate_objective}
\end{equation}
Only one action loss is computed because all four future-state projections correspond to the same physical trajectory and therefore share the same action target. Video padding is handled independently for each projection, while the source-dependent action-dimension mask is retained for Human samples exactly as in CF-WAM. 

\paragraph{Inference.}
At inference time, Static Multi-State jointly denoises four future-state latent streams together with a single action stream. The model receives only the current RGB observation, task conditioning, and proprioceptive state; no ground-truth future depth, segmentation, or DINOv3-PCA frames are provided. Four noisy video latents are initialized independently, while their first temporal position is set to the same encoded current RGB observation. During each denoising step, the action expert reads all four predicted future streams and produces a single action trajectory. The final outputs therefore consist of four predicted future videos and one action sequence. 

\paragraph{Compute-matched comparison with Dynamic-State.}
Static Multi-State and Dynamic-State use the same model initialization, training data, optimizer, learning rate, loss weights, and other optimization settings. However, Static Multi-State simultaneously processes and supervises four future-state streams at every update, resulting in substantially higher per-step computation and slower training than Dynamic-State, which predicts only one sampled projection per training sample. We therefore match the two variants by overall training compute.

Under this compute-matched setting, the comparison evaluates two different ways of using the same set of future-state representations. Static Multi-State allocates computation to all four prediction spaces simultaneously at every update, whereas Dynamic-State supervises one projection at a time and varies the selected projection across updates. The former therefore requires multiple representation-specific objectives to be optimized concurrently within the same action-conditioning model, while the latter allows complementary constraints from different future-state coordinate systems to accumulate across training steps without maintaining all prediction spaces simultaneously.

\begin{table*}[t]
\centering
\caption{
Full evaluation results on the RoboCasa GR1 TableTop benchmark.
We report success rates (\%) over 50 rollouts per task.
}
\label{tab:robocasa_full_results}

\setlength{\tabcolsep}{7.0pt}
\renewcommand{\arraystretch}{1.05}

\resizebox{\textwidth}{!}{%
\begin{tabular}{@{}lrrrrr@{}}
\toprule
\textbf{Task}
& \textbf{ABot-M0}
& \textbf{JoyAI-RA}
& \textbf{FastWAM}
& \textbf{WALA}
& \textbf{CF-WAM} \\
\midrule

CupToDrawerClose
& 48.0 & 48.0 & 32.0 & \textbf{86.0} & 60.0 \\

PotatoToMicrowaveClose
& 50.0 & 70.0 & 72.0 & \textbf{78.0} & 72.0 \\

MilkToMicrowaveClose
& 46.0 & \textbf{84.0} & 68.0 & 78.0 & 72.0 \\

BottleToCabinetClose
& 86.0 & 84.0 & \textbf{92.0} & 82.0 & 86.0 \\

WineToCabinetClose
& 66.0 & 54.0 & 56.0 & 62.0 & \textbf{68.0} \\

CanToDrawerClose
& 74.0 & 90.0 & 84.0 & \textbf{96.0} & 92.0 \\

\cmidrule(lr){1-6}

CuttingboardToBasket
& 70.0 & 88.0 & 84.0 & 86.0 & \textbf{92.0} \\

CuttingboardToCardboardbox
& 58.0 & 46.0 & 64.0 & 66.0 & \textbf{82.0} \\

CuttingboardToPan
& 76.0 & 92.0 & 92.0 & 94.0 & \textbf{100.0} \\

CuttingboardToPot
& 66.0 & 80.0 & 84.0 & 80.0 & \textbf{94.0} \\

CuttingboardToTieredbasket
& 38.0 & 36.0 & 68.0 & 50.0 & \textbf{90.0} \\

\cmidrule(lr){1-6}

PlacematToBasket
& 52.0 & 76.0 & 76.0 & \textbf{96.0} & 88.0 \\

PlacematToBowl
& 66.0 & 52.0 & \textbf{92.0} & 74.0 & 88.0 \\

PlacematToPlate
& 60.0 & 38.0 & 64.0 & 66.0 & \textbf{92.0} \\

PlacematToTieredshelf
& 26.0 & 14.0 & 48.0 & 42.0 & \textbf{50.0} \\

\cmidrule(lr){1-6}

PlateToBowl
& 54.0 & 48.0 & 76.0 & 72.0 & \textbf{86.0} \\

PlateToCardboardbox
& 48.0 & 38.0 & 68.0 & 50.0 & \textbf{82.0} \\

PlateToPan
& 66.0 & 46.0 & 60.0 & 52.0 & \textbf{78.0} \\

PlateToPlate
& 64.0 & 88.0 & 80.0 & 92.0 & \textbf{96.0} \\

\cmidrule(lr){1-6}

TrayToCardboardbox
& 54.0 & 82.0 & 64.0 & 86.0 & \textbf{90.0} \\

TrayToPlate
& 68.0 & 88.0 & 80.0 & \textbf{98.0} & 94.0 \\

TrayToPot
& 64.0 & 88.0 & \textbf{92.0} & 84.0 & 80.0 \\

TrayToTieredbasket
& 60.0 & 62.0 & 88.0 & 80.0 & \textbf{92.0} \\

TrayToTieredshelf
& 38.0 & 24.0 & 44.0 & 54.0 & \textbf{56.0} \\

\midrule

\textbf{Average}
& 58.30
& 63.20
& 72.00
& 75.17
& \textbf{82.50} \\

\bottomrule
\end{tabular}%
}
\end{table*}

\subsubsection{Detailed RoboCasa-GR1 Results}
\label{app:robocasa_detailed}
Table~\ref{tab:robocasa_full_results} reports the per-task success rates of CF-WAM and representative baselines on all 24 RoboCasa-GR1 tabletop tasks, with 50 closed-loop rollouts evaluated for each task. CF-WAM achieves the highest average success rate of \(82.50\%\), outperforming WALA (\(75.17\%\)) and FastWAM (\(72.00\%\)) by \(7.33\) and \(10.50\) percentage points, respectively.

Beyond the overall average, CF-WAM attains the best performance on 15 of the 24 tasks, showing consistent improvements across diverse manipulation settings. The gains are particularly pronounced on several challenging object-placement tasks, such as CuttingboardToTieredbasket (\(90.0\%\)), CuttingboardToCardboardbox (\(82.0\%\)), and PlateToCardboardbox (\(82.0\%\)), where CF-WAM substantially exceeds the strongest competing baseline. At the same time, the per-task results also reveal that no single method dominates every task, highlighting the diversity of manipulation dynamics captured by the benchmark. Overall, the detailed results support the aggregate improvements reported in the main paper and show that CF-WAM's gains are distributed across a broad range of tasks rather than being driven by a small subset.

\subsubsection{Additional RoboCasa-GR1 Rollouts}
\label{app:robocasa_additional_rollouts}
\begin{figure}[t]
    \centering
    \includegraphics[width=\linewidth]{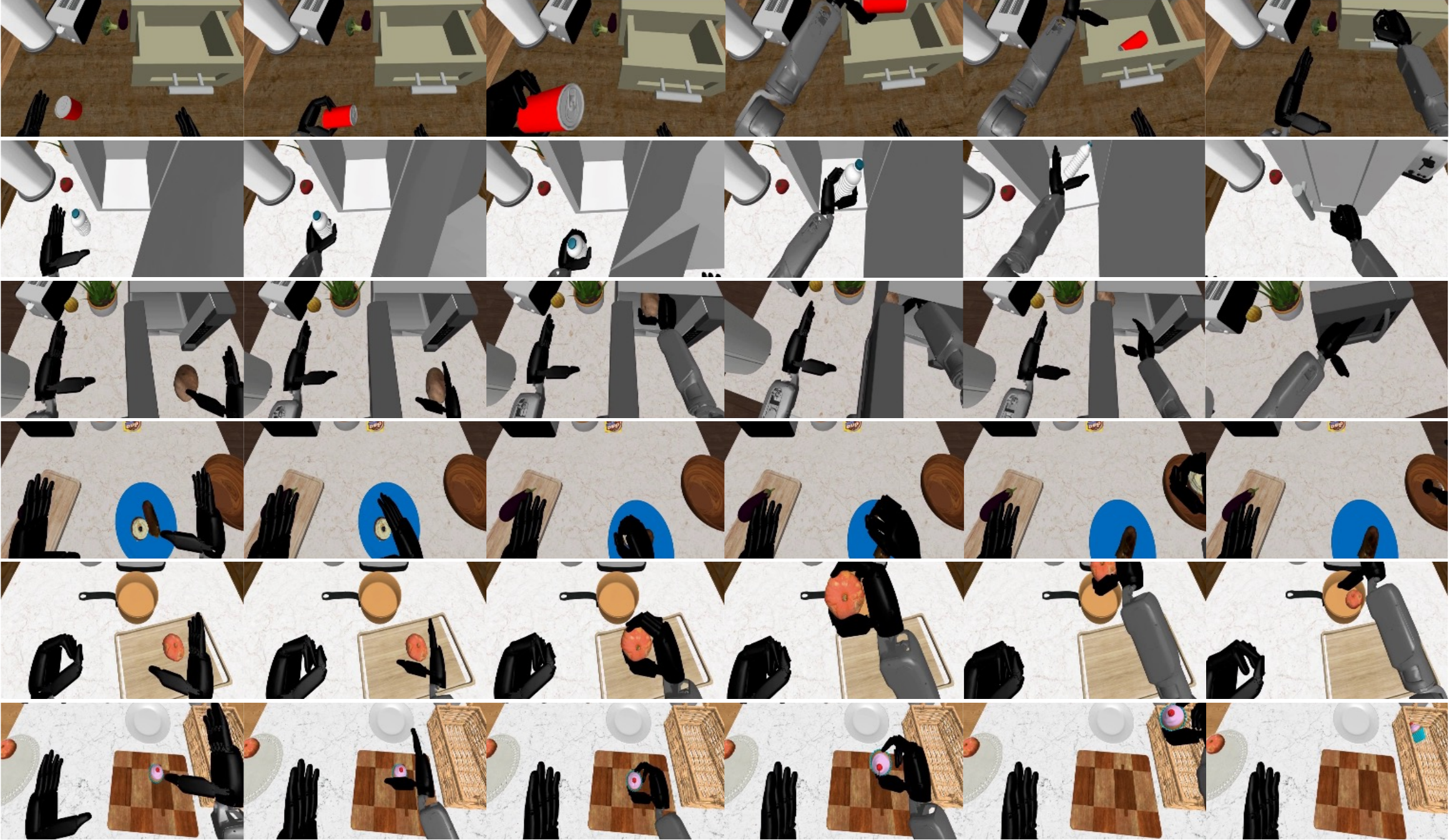}
    \caption{Additional closed-loop manipulation trajectories of CF-WAM on RoboCasa-GR1.}
    \label{fig:robocasa_more_rollouts}
\end{figure}

Figure~\ref{fig:robocasa_more_rollouts} presents additional closed-loop rollouts of CF-WAM across diverse RoboCasa-GR1 manipulation tasks. The trajectories illustrate continuous task progress from the initial observation to successful interaction and object placement, covering different objects, receptacles, viewpoints, and manipulation motions. These qualitative results complement the per-task success rates in Table~\ref{tab:robocasa_full_results} and further demonstrate that CF-WAM can generate coherent action sequences across a broad range of simulated manipulation scenarios.

\subsubsection{Cross-Expert Attention Analysis}
\label{app:attention_analysis}
\begin{figure*}[t]
    \centering
    \includegraphics[width=\textwidth]{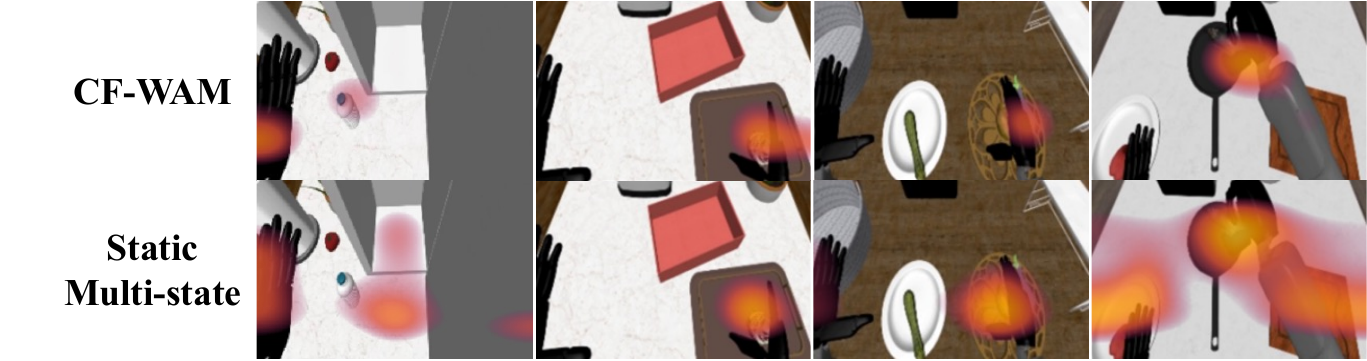}
    \caption{
    Visualization of action-to-video attention for CF-WAM and Static Multi-State across representative RoboCasa-GR1 manipulation scenes.
    }
    \label{fig:attention_comparison}
\end{figure*}

To further examine how different future-state supervision strategies affect action conditioning, we visualize the attention from action queries to video tokens for CF-WAM and Static Multi-State. Figure~\ref{fig:attention_comparison} shows representative RoboCasa-GR1 examples, where warmer regions indicate video locations receiving stronger attention from the action stream.

Across the examples, CF-WAM exhibits more localized attention around manipulation-relevant regions, including the manipulated object, target receptacle, and regions near ongoing contact. In contrast, Static Multi-State produces substantially more diffuse attention, with high-response regions often spreading across multiple objects, manipulators, and surrounding scene areas. This qualitative difference is consistent with the optimization behavior observed in our ablations: simultaneously maintaining four prediction spaces requires the action stream to accommodate representation-specific information from all four futures at every update, which can disperse the limited action-conditioning capacity. Dynamic-State instead exposes one future-state coordinate system at a time, allowing complementary constraints to accumulate across updates while maintaining more focused action-relevant conditioning.

The consistent localization pattern complements the performance gap between Dynamic-State and Static Multi-State reported in the main paper.

\subsubsection{Human--Robot Action-Space Ablations}
\label{app:action_space_ablation}
Table~\ref{tab:action_space_ablation} examines how different Human--Robot action-space designs affect policy learning. When both Human and Robot trajectories are represented using only joint actions, the model achieves an average success rate of \(76.80\%\). Providing both domains with the full EE+Joint representation does not improve performance, yielding \(75.00\%\). In contrast, our Human--Robot design, which represents Human actions through the shared end-effector space while retaining EE+Joint control for Robot trajectories and masking Robot-specific dimensions for Human samples, improves the average success rate to \(78.25\%\).

These results suggest that simply increasing the action representation shared across Human and Robot data is not sufficient. Instead, preserving the action dimensions that are naturally shared across embodiments while retaining Robot-specific control dimensions provides a more effective interface for Human--Robot learning. Finally, applying Dynamic-State supervision on top of this action-space design further increases performance from \(78.25\%\) to \(82.50\%\). This \(4.25\)-point improvement shows that the gains from dynamic future-state supervision are complementary to those obtained from the Human--Robot action-space design.

\begin{table}[t]
    \centering
    \caption{
    Ablation of Human--Robot action-space designs on RoboCasa-GR1.
    Success rates (\%) are reported across 24 tasks.
    }
    \label{tab:action_space_ablation}

    \setlength{\tabcolsep}{5pt}
    \renewcommand{\arraystretch}{1.05}

    \begin{tabular}{llllc}
        \toprule
        Setting
        & Human Action
        & Robot Action
        & Future State
        & Avg. \\
        \midrule

        Joint-only
        & Joint
        & Joint
        & Visual
        & 76.8 \\

        EE+Joint
        & EE + Joint
        & EE + Joint
        & Visual
        & 75 \\

        Human--Robot
        & EE
        & EE + Joint
        & Visual
        & 78.25 \\

        \midrule

        \textbf{CF-WAM}
        & EE
        & EE + Joint
        & Dynamic
        & \textbf{82.50} \\

        \bottomrule
    \end{tabular}
\end{table}

\subsubsection{Real-World Tasks and OOD Settings}
\label{app:realworld_setup}

\paragraph{Real-world manipulation tasks.}
We evaluate CF-WAM on six real-world manipulation tasks that cover a diverse set of interaction patterns, including rigid-object transport, deformable-object manipulation, object organization, semantic sorting, precise insertion, and contact-rich surface wiping. Each task is evaluated with 50 closed-loop trials.

\textbf{Chemistry.}
The robot is required to grasp an Erlenmeyer flask containing liquid and place it onto a weighing platform. This task requires stable grasping and transport of a partially filled container while accurately reaching a spatially constrained target region.

\textbf{Folding.}
The robot folds a towel placed on the tabletop. Unlike rigid-object manipulation, this task involves deformable-object dynamics and requires the policy to coordinate contact locations and motion over an extended spatial region.

\textbf{Organization.}
The robot organizes the workspace by picking up a medicine bottle from the tabletop and placing it into a drawer. The task combines object grasping, spatial relocation, and interaction with a storage region that partially occludes the target during execution.

\textbf{Fruit Sorting.}
The robot performs fruit sorting by identifying and moving a fruit to its corresponding target region. This task evaluates whether the policy can couple semantic object understanding with goal-directed manipulation under variations in object appearance and initial placement.

\textbf{Insertion.}
The robot grasps a test tube and inserts it back into a designated hole of a test-tube rack. The small insertion tolerance makes this task particularly sensitive to end-effector positioning and orientation, providing a test of fine-grained manipulation accuracy.

\textbf{Wiping.}
We create a visible tabletop stain using instant coffee and require the robot to wipe the contaminated region. This task involves sustained surface contact and spatially extended motion, rather than a single pick-and-place interaction, and therefore evaluates contact-rich manipulation and trajectory-level control.

\paragraph{Evaluation protocol.}
All real-world experiments are conducted in closed loop. For each task and each evaluated setting, we perform 50 independent trials and report the corresponding task success rate. A trial is counted as successful only when the intended task objective is completed at the end of execution.

\paragraph{Human-like and OOD settings.}
In addition to the standard real-world evaluation, we evaluate the policy under five out-of-distribution settings that perturb different aspects of the test distribution. These settings include a Human-like scenario and four controlled visual or spatial variations.

\textbf{Human-like.}
We construct a real-world manipulation scenario whose scene configuration is similar to those observed in the Human video data but is not covered by the Robot training data. This setting evaluates whether incorporating Human demonstrations can improve Robot generalization to manipulation contexts that are absent from the Robot data distribution but related to experience available from the Human domain.

\textbf{Background texture.}
We alter the visual texture or appearance of the workspace background while keeping the task semantics unchanged, testing robustness to changes in scene appearance.

\textbf{Lighting.}
We modify the illumination conditions of the workspace, introducing changes in brightness, shading, and visual contrast without altering the physical task.

\textbf{Object instance.}
We replace task objects with different instances that preserve the same functional role. This setting evaluates whether the learned policy can generalize beyond the specific object appearances observed during training.

\textbf{Object position.}
We perturb the initial positions of task-relevant objects while keeping their functional roles and the task goal unchanged. This setting evaluates robustness to spatial rearrangements of the manipulation scene.

\subsubsection{Real-World OOD Results}
\label{app:realworld_ood}
Table~\ref{tab:realworld_ood} reports the real-world OOD performance of CF-WAM under different inference projections. Across the five OOD settings, all four projections maintain strong performance, with average success rates ranging from \(73.20\%\) to \(79.20\%\). Interaction achieves the highest overall average of \(79.20\%\), followed by Geometric (\(76.00\%\)), Visual (\(75.60\%\)), and Semantic (\(73.20\%\)). The relative advantages vary across OOD factors: Interaction performs best under background-texture and lighting changes, while Visual and Geometric are particularly robust to object-position shifts. This variation further suggests that different future-state projections capture complementary information useful under different distribution shifts.

Human experience plays an especially important role in OOD generalization. With Visual inference, removing Human data reduces the average success rate from \(75.60\%\) to \(45.60\%\), a drop of \(30.00\) percentage points. The largest gap appears in the Human-like setting, where performance decreases from \(82.00\%\) to \(38.00\%\). This result is consistent with the construction of the Human-like evaluation: although these scenes are absent from the Robot training distribution, related scene configurations are present in the Human data. Overall, the results indicate that Human experience substantially broadens the generalization range of the Robot policy, while the different future-state projections provide complementary robustness to visual, object, and spatial distribution shifts.

\begin{table}[t]
\centering
\caption{
Real-world OOD analysis under different inference projections and Human experience.
We report success rates (\%) under five OOD settings and their average.
}
\label{tab:realworld_ood}

\setlength{\tabcolsep}{3.0pt}
\renewcommand{\arraystretch}{1.05}

\begin{tabular}{@{}llcccccc@{}}
\toprule
Setting
& Inference
& Human-like
& Bg. Texture
& Lighting
& Obj. Inst.
& Obj. Pos.
& Avg. \\
\midrule

\multirow{4}{*}{CF-WAM}
& Visual
& \textbf{82.00} & 64.00 & 74.00 & \textbf{70.00} & \textbf{88.00} & 75.60 \\

& Semantic
& 74.00 & 70.00 & 74.00 & 64.00 & 84.00 & 73.20 \\

& Geometric
& 76.00 & 78.00 & 76.00 & 62.00 & \textbf{88.00} & 76.00 \\

& Interaction
& 80.00 & \textbf{82.00} & \textbf{80.00} & 68.00 & 86.00 & \textbf{79.20} \\

\cmidrule(lr){1-8}

w/o Human
& Visual
& 38.00 & 42.00 & 50.00 & 36.00 & 62.00 & 45.60 \\

\bottomrule
\end{tabular}
\end{table}

\subsubsection{Additional Real-World Rollouts}
\label{app:realworld_additional_rollouts}
\begin{figure}[t]
    \centering
    \includegraphics[width=\linewidth]{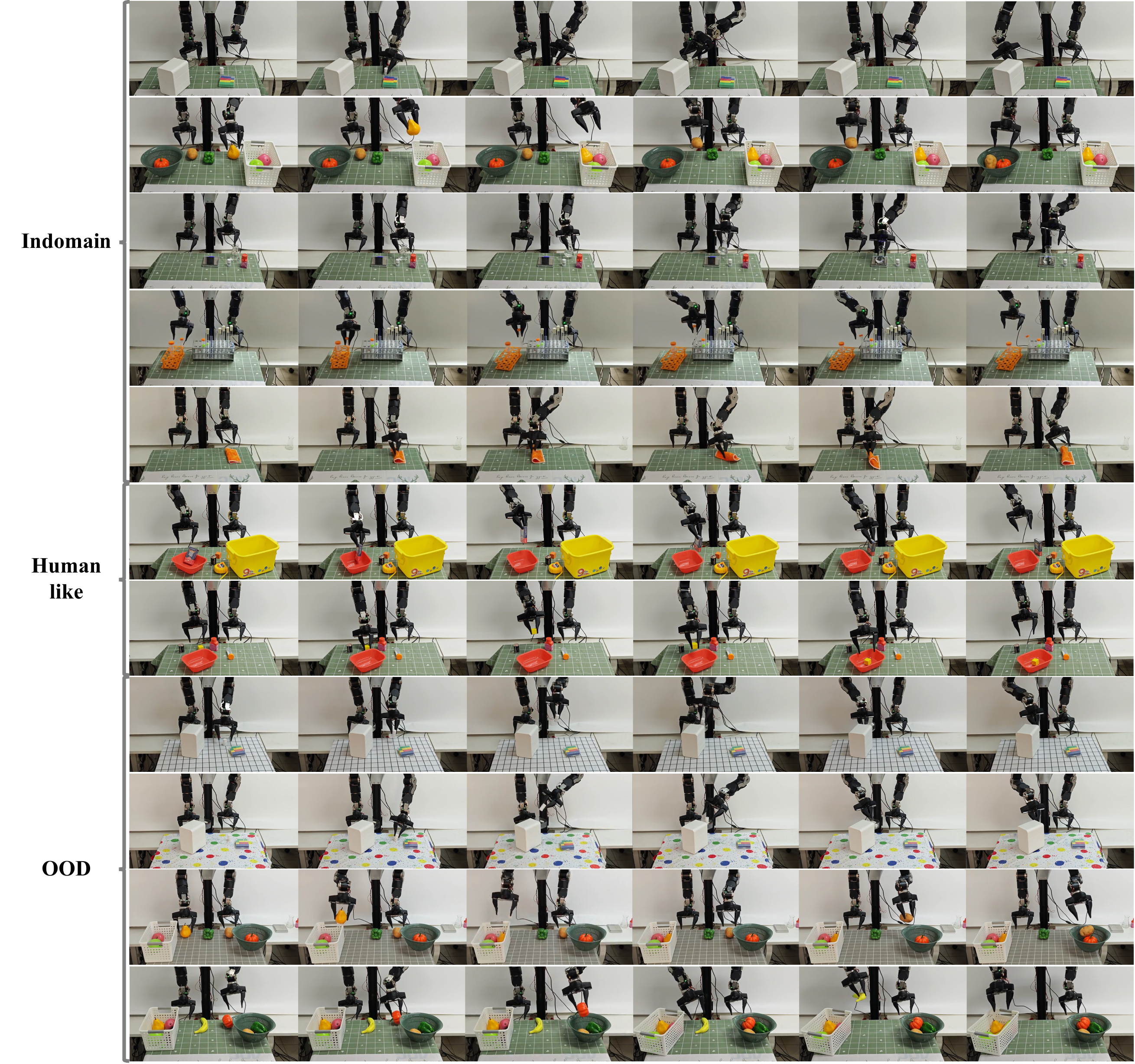}
    \caption{Additional real-world manipulation trajectories of CF-WAM under in-domain, Human-like, and OOD settings.}
    \label{fig:realworld_more_rollouts}
\end{figure}

Figure~\ref{fig:realworld_more_rollouts} presents additional closed-loop real-world trajectories of CF-WAM under in-domain, Human-like, and OOD settings. The examples cover diverse manipulation behaviors, including object transport, sorting, insertion, and other contact-rich interactions. Across in-domain scenes, CF-WAM produces coherent action sequences that consistently progress toward task completion.

The Human-like examples further show successful execution in scene configurations that are not covered by the Robot training data but are related to experience available from the Human domain. Under OOD conditions, including changes in background appearance, lighting, object instances, and object positions, the policy continues to maintain task-directed behavior despite substantial visual and spatial variations. These qualitative results complement the quantitative analysis in Table~\ref{tab:realworld_ood}, illustrating that the generalization gains of CF-WAM extend across both Human-related scene shifts and controlled OOD perturbations.

\end{document}